\documentclass[letterpaper, 10 pt, conference]{ieeeconf}  % Comment this line out if you need a4paper

\IEEEoverridecommandlockouts                              % This command is only needed if 
\usepackage{graphics} % for pdf, bitmapped graphics files
\usepackage{graphicx}
\usepackage[svgnames]{xcolor}

\usepackage{amsmath} % assumes amsmath package installed
\usepackage{booktabs} 
\usepackage{balance}

\usepackage{subcaption}     % for subfigure captions
\newif\ifanonymous
\anonymousfalse

\title{\LARGE \bf
HACL: History-Aware Curriculum Learning for Fast Locomotion
}

\ifanonymous
  \author{Anonymous}
\else
  \author{
  Prakhar Mishra$^{1,*}$,
  Amir Hossain Raj$^{2}$,
  Xuesu Xiao$^{2}$,
  and Dinesh Manocha$^{1}$%
  \thanks{$^{1}$University of Maryland, College Park, MD, USA.}%
  \thanks{$^{2}$George Mason University, Fairfax, VA, USA.}%
  \thanks{*Collaborating Researcher; M.Eng. University of Maryland, 2023.}%
  }
\fi

\begin{document}

\maketitle
\thispagestyle{empty}
\pagestyle{empty}

%%%%%%%%%%%%%%%%%%%%%%%%%%%%%%%%%%%%%%%%%%%%%%%%%%%%%%%%%%%%%%%%%%%%%%%%%%%%%%%%
\begin{abstract}

% We address the problem of agile and rapid locomotion, a key characteristic of quadrupedal and bipedal robots. We present a new algorithm that maintains stability and generates high-speed trajectories by incorporating the temporal dynamics of locomotion. 
% %Our approach is designed to improve the performance of current methods based on fixed rules and multi-terrain characteristics by considering the temporal aspect of locomotion. 
High-speed legged locomotion struggles with stability at higher command velocities as most curriculum approaches advance task difficulty aggressively. We address this by adapting our sampler based on reward history instead of instantaneous returns or fixed update rules.
Our formulation integrates past information via a novel history-aware curriculum learning (HACL) algorithm. We model the history of velocity commands with respect to the observed linear and angular rewards with a recurrent neural net (RNN) using a History-Aware Greedy (HA-Greedy) sampling procedure.
%In our results and validation sections, we demonstrate the importance of a temporal modeling-based curriculum approach, which is missing from the current curriculum learning and RL methods.
We validate our approach on the MIT Mini Cheetah and Unitree Go1 and Go2 robots in a simulated environment trained independently to demonstrate morphology generalization. Our HACL policy achieves a velocity of 6.7 m/s in simulation and outperforms prior locomotion algorithms \cite{margolis2024rapid}, \cite{aractingi2023controlling}. Finally, we deploy our HACL policy on a Unitree Go1 robot across various real-world terrains (pebbles, slopes, tile, concrete floor, grass, etc.), achieving a velocity of 4.1 m/s. To our knowledge, it is among the highest reported RL-based controllers under comparable settings \cite{margolis2024rapid}, \cite{kumar2021rma}, \cite{rudin2022advanced}, \cite{hwangbo2019learning}, \cite{aractingi2023controlling}. 

\end{abstract}

%%%%%%%%%%%%%%%%%%%%%%%%%%%%%%%%%%%%%%%%%%%%%%%%%%%%%%%%%%%%%%%%%%%%%%%%%%%%%%%%
\section{INTRODUCTION}

Agility and speed are essential for legged robots in unstructured environments, where terrain diversity affects both performance and stability. Difficulties arise as robots often lack accurate knowledge of the real-world properties such as unknown friction, obstacles, uneven terrain, slippery or inclined surfaces, etc. Model-based and learning-based methods attempt to model the environment parameters \cite{zucker2010optimization}, \cite{hwangbo2019learning}, \cite{yin2007simbicon} to improve locomotion, but designing robust model-based controllers requires substantial human expertise and may still not capture the full dynamics. Recent reinforcement learning and imitation learning-based methods reduce this dependence on human expertise in modeling dynamics and have been shown to improve robot performance \cite{hwangbo2019learning}, \cite{peng2020learning}. However, training such RL methods poses additional challenges for rapid locomotion: sparse rewards and underexploration can lead to instability or failure at higher command velocities \cite{margolis2024rapid}, \cite{rudin2022advanced}, \cite{xie2020allsteps}. When policies are trained with higher command velocity, they often fail to generalize compared to when trained in a narrow range with gradual increase; it is similar to multitask reinforcement learning, where the policy may myopically start optimizing immediate or easier tasks instead of acquiring generality between tasks \cite{hessel2019multi}. 

\begin{figure}[h]
    \centering

    % Row 1: Pebbles
    \begin{minipage}{0.32\linewidth}
        \centering
        \includegraphics[width=\linewidth]{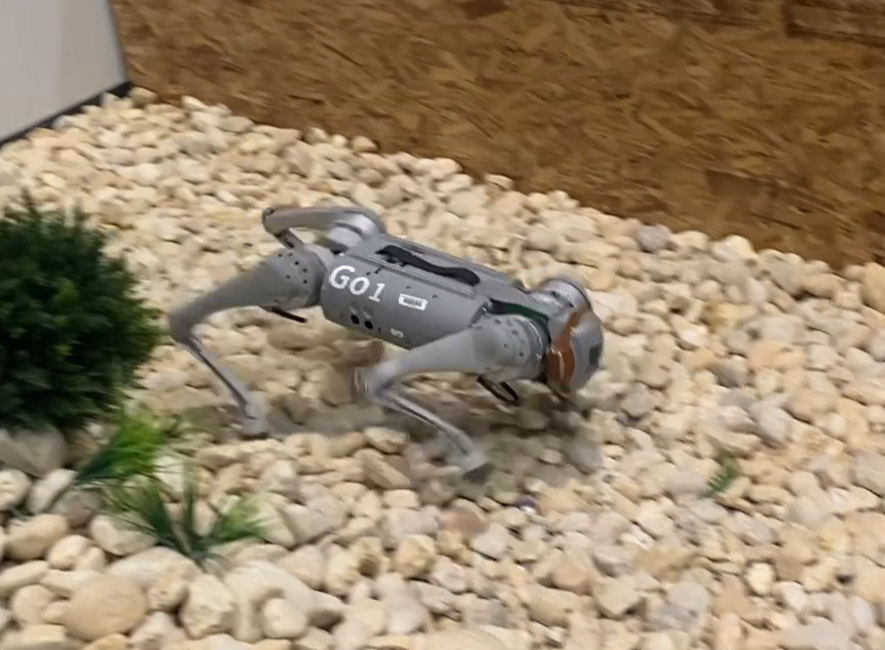}
    \end{minipage}
    \begin{minipage}{0.32\linewidth}
        \centering
        \includegraphics[width=\linewidth]{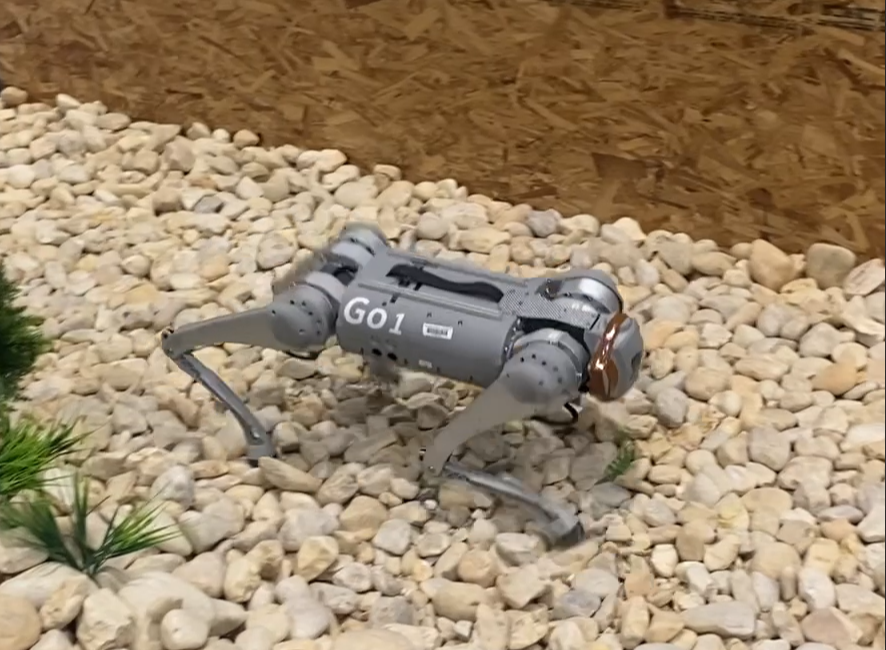}
    \end{minipage}
    \begin{minipage}{0.32\linewidth}
        \centering
        \includegraphics[width=\linewidth]{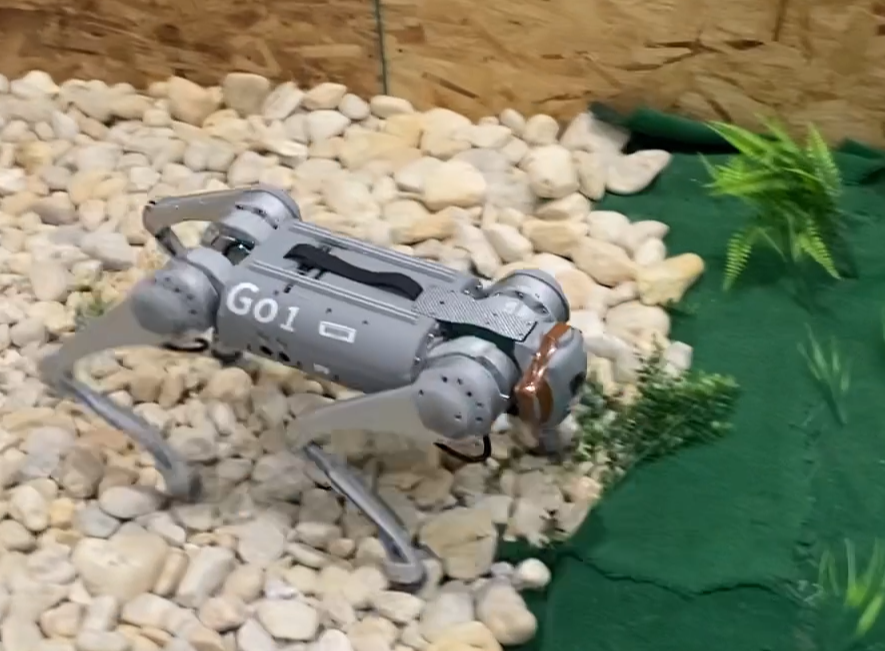}
    \end{minipage}
    
    % Row 2: Inclined slopes
    \begin{minipage}{0.32\linewidth}
        \centering
        \includegraphics[width=\linewidth]{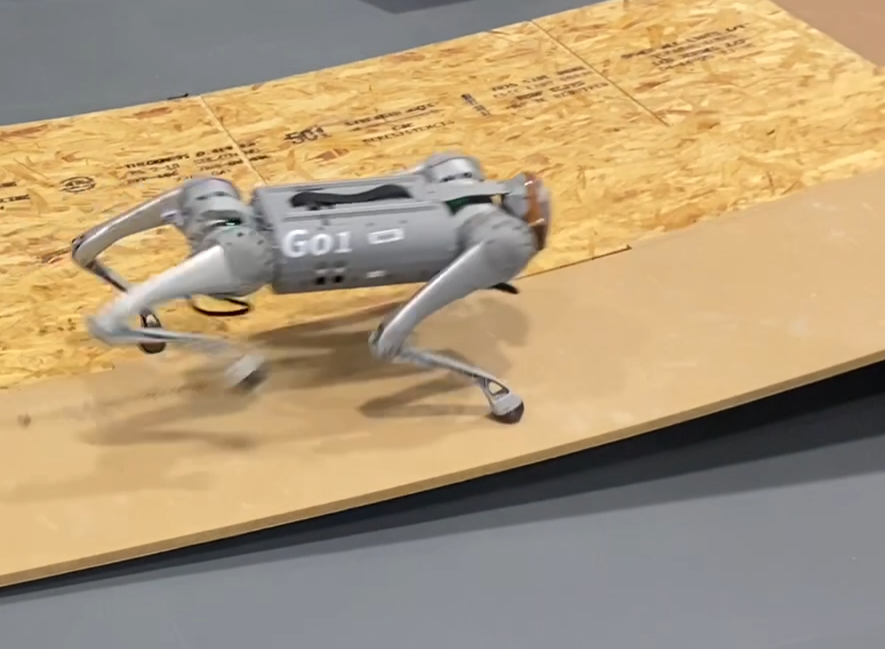}
    \end{minipage}
    \begin{minipage}{0.32\linewidth}
        \centering
        \includegraphics[width=\linewidth]{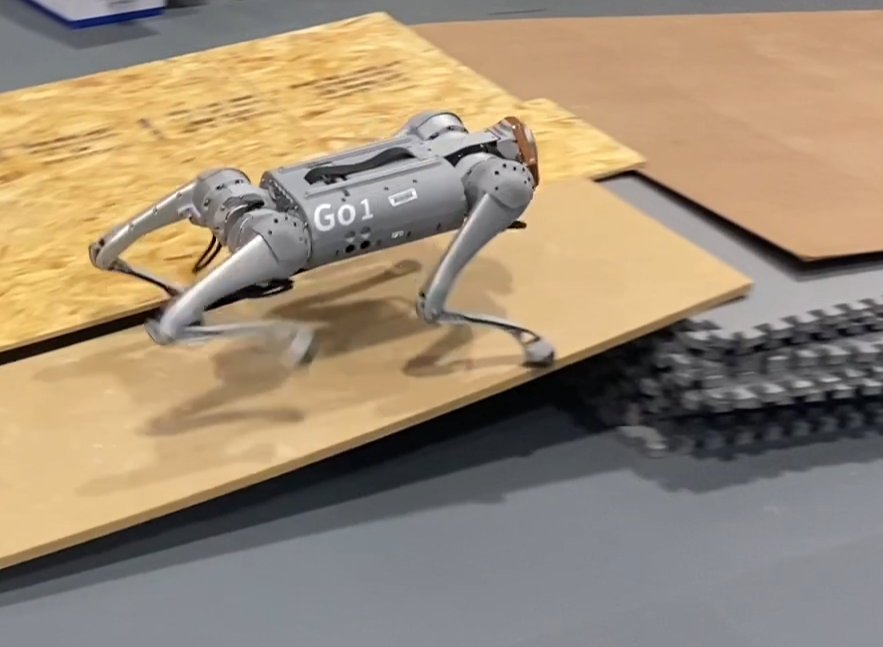}
    \end{minipage}
    \begin{minipage}{0.32\linewidth}
        \centering
        \includegraphics[width=\linewidth]{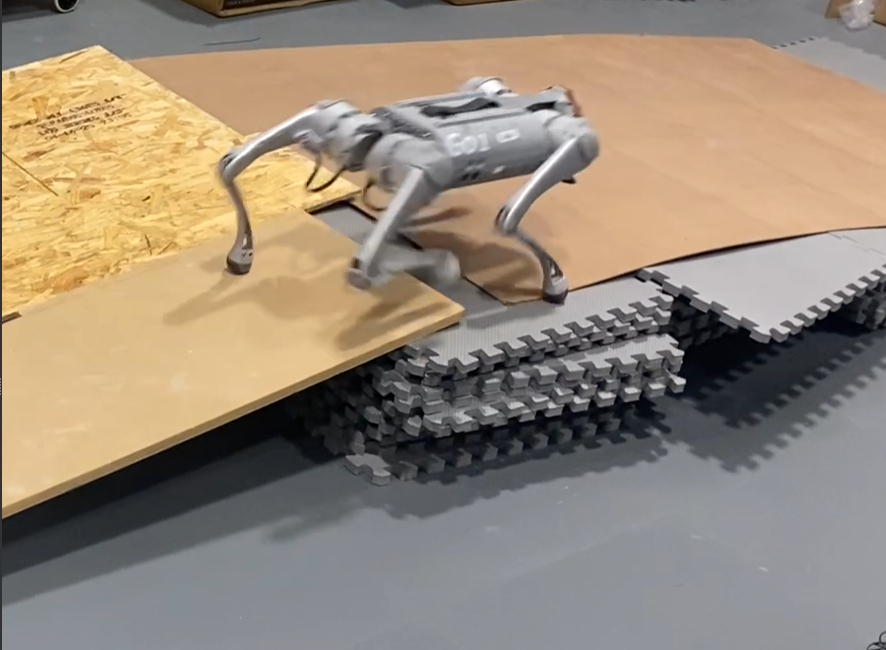}
    \end{minipage}

    % Row 3: Broken rocks
    \begin{minipage}{0.32\linewidth}
        \centering
        \includegraphics[width=\linewidth]{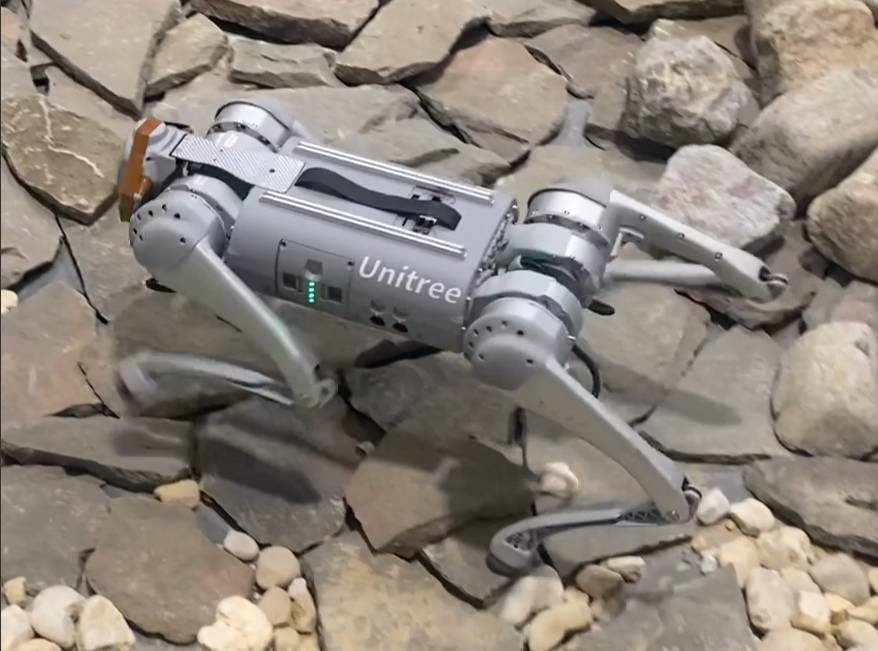}
    \end{minipage}
    \begin{minipage}{0.32\linewidth}
        \centering
        \includegraphics[width=\linewidth]{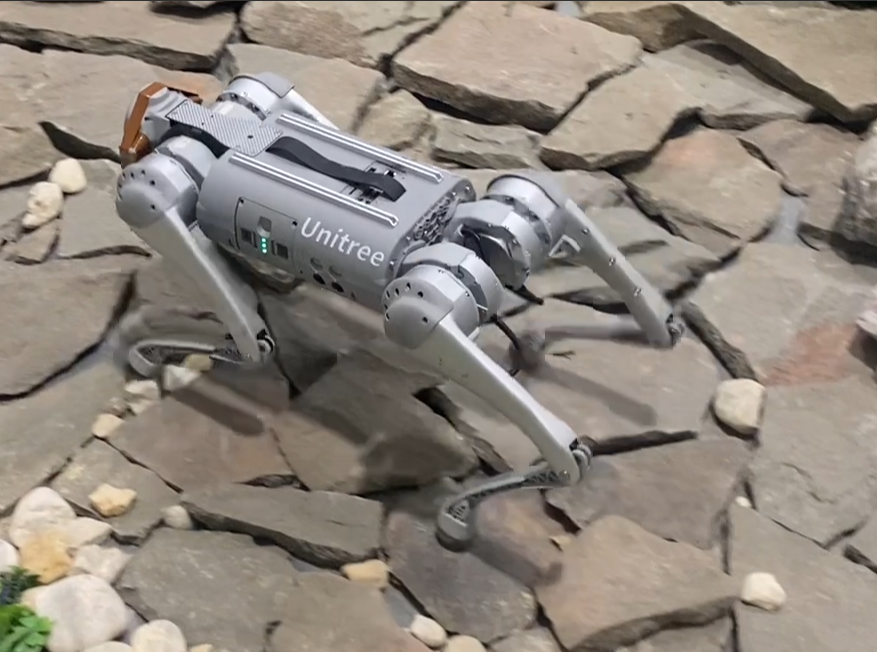}
    \end{minipage}
    \begin{minipage}{0.32\linewidth}
        \centering
        \includegraphics[width=\linewidth]{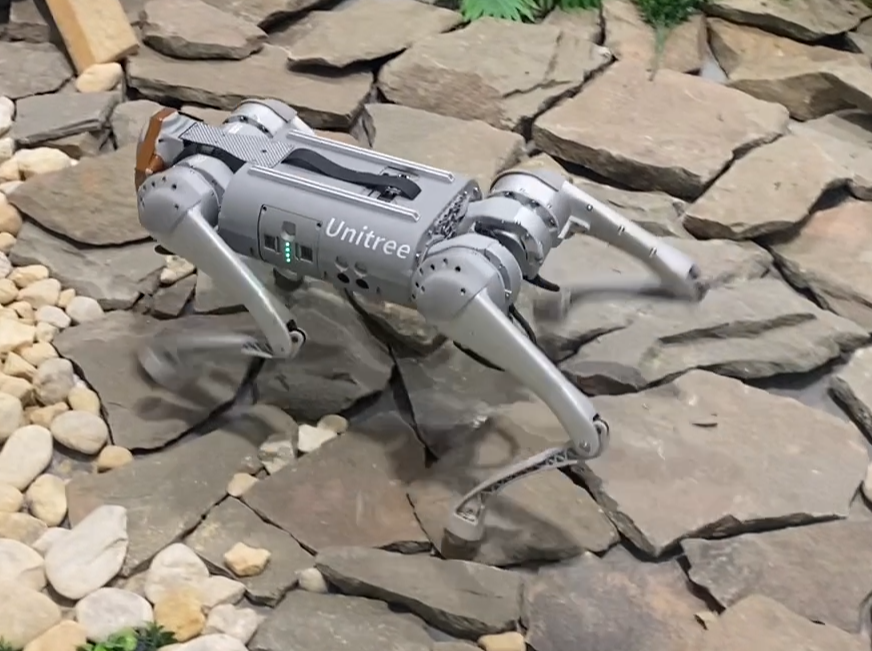}
    \end{minipage}

    % Row 4: Angular velocity rotation
    \begin{minipage}{0.32\linewidth}
        \centering
        \includegraphics[width=\linewidth]{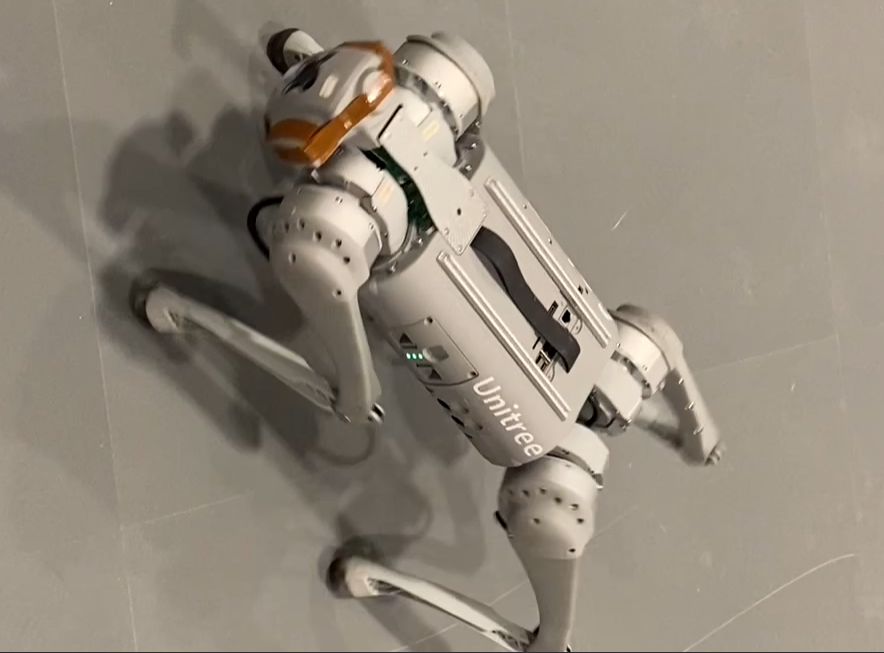}
    \end{minipage}
    \begin{minipage}{0.32\linewidth}
        \centering
        \includegraphics[width=\linewidth]{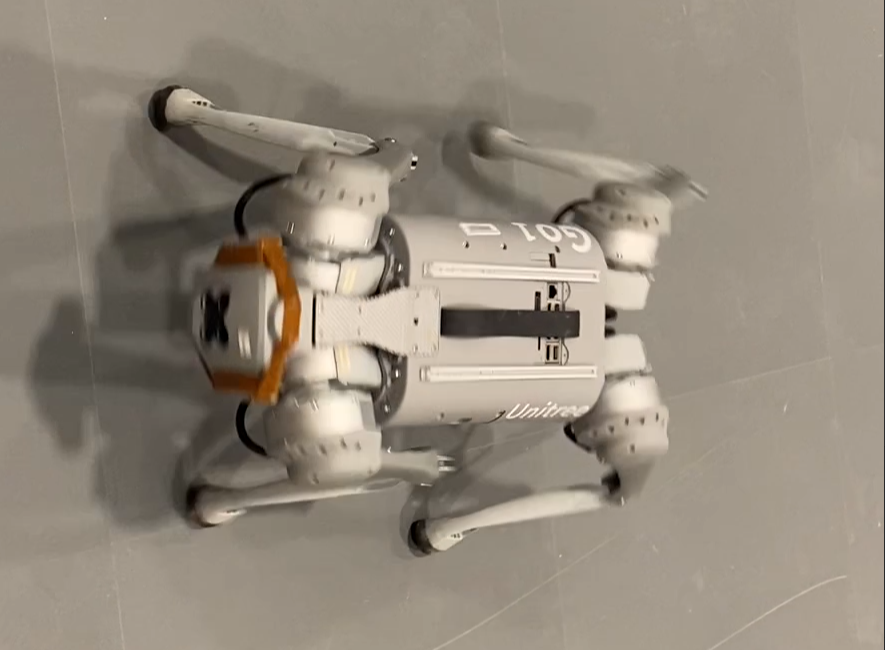}
    \end{minipage}
    \begin{minipage}{0.32\linewidth}
        \centering
        \includegraphics[width=\linewidth]{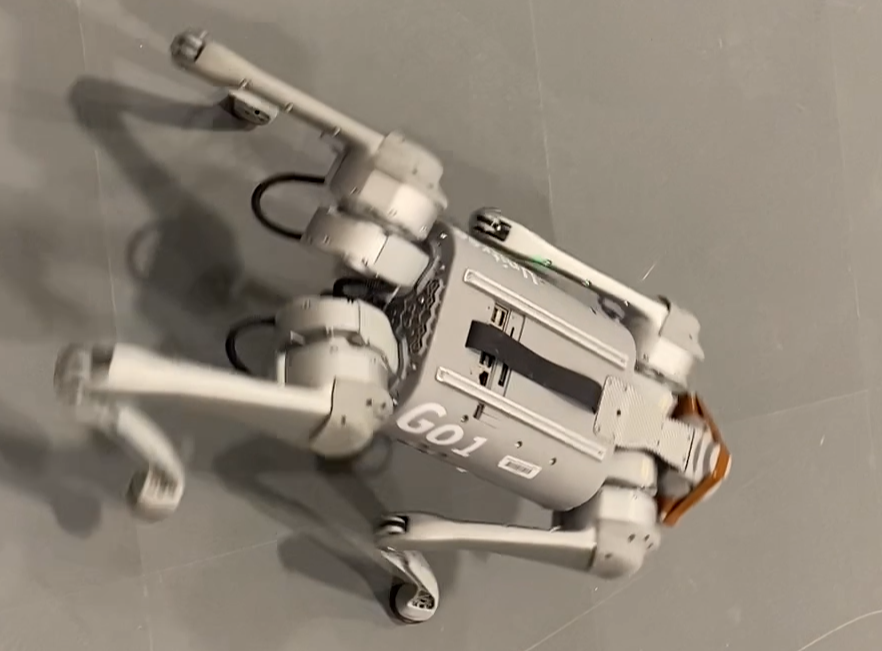}
    \end{minipage}

    \caption{\textbf{Testing on diverse terrains (Unitree Go1, deployed HACL).}      
    \textbf{Row 1:} On Pebbles (2-3 m) Go1 maintains $2.1 \pm 0.3$ m/s with a success rate of 60\% and a lateral shift of $0.3 \pm 0.1$ m. 
    \textbf{Row 2:} On wooden slopes (approximate $20^\circ$ and 3-5 m), the robot achieves $3.1 \pm 0.4$ m/s with a success rate of 80\% and a lateral shift of $0.5 \pm 0.3$ m.
    \textbf{Row 3:} 2-3 m run; $1.5 \pm 0.4$ m/s ; the success rate hovers around 50\% and the performance is very similar to Pebbles.
    \textbf{Row 4:} For angular rotation it achieves $3.7 \pm 0.2$ rad/s and a success rate of 100\% for the given angular command velocity. HACL maintains high speed and stability across varied tasks and terrains.}
    \label{fig:hardware_robust}
\end{figure}

One way to address high velocity locomotion challenges is to implement curriculum learning \cite{bengio2009curriculum} in locomotion tasks \cite{margolis2024rapid}, \cite{kumar2021rma}, \cite{rudin2022learning}. Control over task difficulty, sample efficiency, and policy optimization are the key reasons curriculum learning has been explored in robotic contexts \cite{li2020towards}, \cite{matiisen2019teacher}. Given its utility, curriculum learning has gained traction in locomotion including walking, running, trotting, pacing, turning, climbing slope, etc. \cite{margolis2023walk}, \cite{margolis2024rapid}, \cite{lee2020learning}, \cite{zhang2024learning}.  

However, many automatic curricula like Rapid-Motor Adaptation (RMA) \cite{kumar2021rma}, Solo12 \cite{aractingi2023controlling}, Gaussian Mixture Models (ALP-GMM) \cite{portelas2020teacher}, etc. prioritize sample efficiency but ignore temporal dependencies, limiting gait optimization. Existing works \cite{margolis2024rapid}, \cite{rudin2022learning}, \cite{hwangbo2019learning}, \cite{zhang2024learning}, \cite{aractingi2023controlling}, \cite{kumar2021rma}, use curriculum for locomotion, but typically ignore the temporal evolution of rewards during the bin selection process. Ignoring these dependencies hinders modeling rewards evolution across episodes: the policy might be overfit to immediate high-reward bins while neglecting bins with higher expected cumulative return. Consequently, such methods can reach basic competency but often struggle to maintain very high velocity and stability.

To address this, rather than feeding reward history as an input to the actor-critic policy, we exploit it at the meta- (curriculum) level. Our History-Aware Curriculum Learning (HACL) integrates the temporal structure into the curriculum, reducing hand-tuning of schedules while explicitly balancing agility and stability (see results). Importantly, HACL generalizes across MIT Mini Cheetah, Unitree Go1, and Go2 in sim and on real Go1; specific details in Section V.

\noindent {\bf Main Results:} We introduce \emph{History-Aware Curriculum Learning (HACL)}, a curriculum strategy that models temporal dependencies by sampling the training history leading to fast and stable locomotion. Unlike previous curriculum strategies which inherently consider locomotion as Markovian ~\cite{byl2008metastable}, \cite{zhao2020non}, \cite{yu2019sim}, HACL leverages the history of task reward to guide the curriculum towards higher reward bins, reducing inefficient exploration. The key contributions of our work include:

\begin{itemize}
    \item \textbf{Method.} HACL influences command bins based on predicted reward utilizing history (\emph{HA-Greedy}), reducing inefficient curriculum updates compared to other baselines \cite{margolis2024rapid}, \cite{rudin2022advanced}, \cite{aractingi2023controlling}. (Section III)
% A History-Aware Curriculum Learning (HACL) that schedules the next episode bins based on previous observed linear and angular velocity rewards $(r_{lin}, r_{ang})$. Our \emph{HA-Greedy} sampling rule exploits the temporal structure, which is generally ignored by fixed curriculum strategies\cite{margolis2024rapid}, \cite{rudin2022advanced}, \cite{aractingi2023controlling}. (Section III)
    \item \textbf{Approach.} By utilizing the hidden layers ($h_{t-1}$) of recurrent neural networks, we bridge the gap between non-Markovian dynamics and curriculum learning,  conditioning on high rewards bins and accelerating convergence.  (Section III)
    \item \textbf{Evaluation.} We extensively evaluated HACL in simulation on the MIT Mini Cheetah and Unitree Go1 and Go2 robots in an IsaacGym simulator, demonstrating generalization across quadrupedal morphologies. And finally validated it on the Unitree Go1 robot in the real-world conditions across various terrains.(Section III-V)
    \item \textbf{Outcome.} HACL achieves velocity of $6.7 \pm 0.2$ m /s in simulation and $4.1 \pm 0.2$ m/s on Go1 Hardware with yaw of $3.7 \pm 0.2$ rad/s, among the fastest RL-Based controller to our knowledge \cite{margolis2024rapid}, \cite{kumar2021rma},\cite{aractingi2023controlling} ,\cite{rudin2022learning}. HACL improves stability \((S\uparrow\ 2000)\) and reduces energy consumption per distance \((CoT\downarrow\!2.6)\) and outperforms fixed curricula, bandit-based task sampling, and non-history networks (Section IV-V).
    
\end{itemize}

\begin{figure}[thpb]
  \centering     
  \includegraphics[width=\columnwidth]{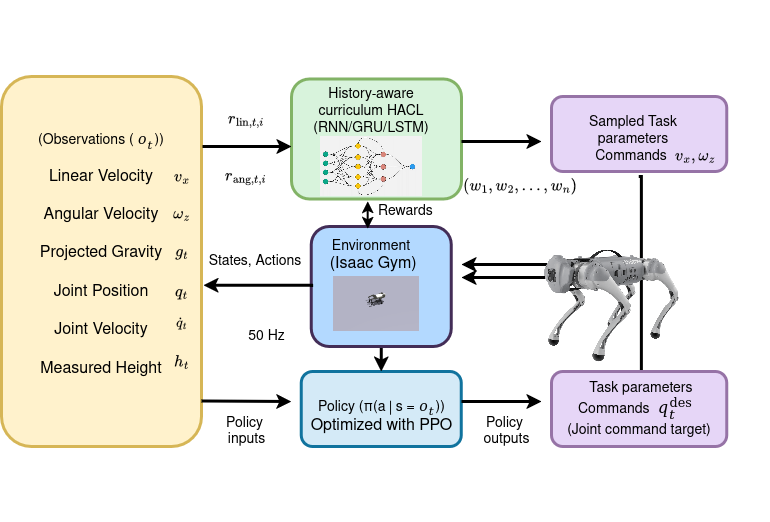}
  \caption{\textbf{HACL overview.} Our \textcolor{ForestGreen}{HACL module} receives the \textcolor{Goldenrod}{observations} and rewards from the \textcolor{blue}{IsaacGym environment}. HACL learns the hidden pattern between these reward distributions and the high-level sampled \textcolor{DeepPink}{task parameters}. The \textcolor{Indigo}{policy model} is optimized using \textbf{PPO} and generates the low-level task parameter commands, which the \textcolor{DimGray}{Unitree Go1} executes in the environment.}
 \label{fig:hacl_overview}
\end{figure}

\section{Related Work}

\subsection{Curriculum Learning in Robotics} Curriculum learning has gained significant traction in robotics research. \cite{bengio2009curriculum} introduced the continuation method style curriculum,and \cite{wang2021survey} surveys the predefined curriculum and the automatic curriculum by grouping them into self-paced, RL Teacher \cite{matiisen2019teacher}, and Other Automatic CL. For locomotion, multi-terrain or environment progression-based curriculum have been used \cite{aractingi2023controlling}, \cite{kumar2021rma}, along with hindsight \cite{chen2024reinforcement}, arm wheel curriculum \cite{wang2024arm}, and guided-terrain curriculum \cite{tidd2020guided}. Fixed rules-based schedules can learn locomotion control policy \cite{margolis2024rapid}, but none exploit the reward history like HACL.

% ,\cite{li2024learning}

% As \cite{bengio2009curriculum} proposes a method similar to continuation methods, \cite{wang2021survey} discusses how to implement / design a predefined curriculum or an automatic curriculum. Their work categorizes automatic curriculum learning (CL) into 4 categories: Self-paced learning, RL Teacher~\cite{matiisen2019teacher}, and Other Automatic CL. While \cite{aractingi2023controlling} employs a multi-terrain curriculum approach and similarly environment progression based on curriculum \cite{kumar2021rma} for learning locomotion skills. And \cite{chen2024reinforcement} ,\cite{li2024learning} discuss hindsight curriculum learning, \cite{wang2024arm} arm wheel curriculum, and \cite{tidd2020guided} guided curriculum for terrains.
% Although \cite{margolis2024rapid}, used a fixed rule-based curriculum update method to learn locomotion control policy, the lack of historical information was common in all these works.

\subsection{Non-Markovian Nature of Locomotion} 

Locomotion exhibits a non-Markovian nature, where the state depends not only on
immediate control inputs, but also on the previous states \cite{byl2008metastable,zhao2020non,yu2019sim}. Previous work addressed this through state-space re-planning and robustness techniques or by discrete non-Markov chain \cite{byl2008metastable}, \cite{defazio2024learning}. Hindsight curricula relax the Markovian nature \cite{li2024learning}. This motivates us to explore history-aware methods.

% As \cite{byl2008metastable,zhao2020non,yu2019sim} have pointed out, locomotion has a non-Markovian nature, where the state depends not only on immediate control inputs, but also on the previous states~\cite{zhao2020non}. These researchers have used state space re-planning strategies and robust bundles, implicitly acknowledging the importance of history. While \cite{byl2008metastable} suggests a discrete Markov chain or a non-Markovian chain, \cite{defazio2024learning} suggests Markovian challenges. \cite{li2024learning} hindsight curriculum learning also addresses Markov property and \cite{george2021learning} implies the Markovian nature of locomotion.

\subsection{Neural Network Architectures and Bandits} 
% , \cite{rodriguez2021deepwalk}

Neural networks are widely used for legged locomotion \cite{zhao2020non}. Bandit schedulers (e.g., UCB \cite{auer2002using} and Thompson sampling \cite{chapelle2011empirical}) have not been used in quadrupedal scenarios, but other robotic scenarios \cite{pini2012multi}, and self-paced learning \cite{kasaei2020learning} adapts to task difficulty. However, these methods do not model the reward history to guide the core curriculum of our HACL.

% Previous works \cite{zhao2020non} have used neural net for legged locomotion, while \cite{rodriguez2021deepwalk} used a neural net to capture hidden state information. Some methods like UCB~\cite{auer2002using}, Thompson sampling~\cite{thompson1933likelihood} which are good at balancing exploration and exploitation, have not been used in quadrupedal but other robotic scenarios \cite{pini2012multi}. And methods like Self-Paced learning \cite{kasaei2020learning} explore adaptive curriculum to balance exploration and exploitation. However, these methods have not been used to learn the reward pattern over a period of time like HACL.

\section{History-Aware Curriculum Learning (HACL)}
Quadrupedal robots struggle to sustain higher velocities as the command-velocity range widens. Standard curricula typically ignore the history of rewards, which can capture latent dynamics (e.g., sensor latency and actuator delays). We formulate the History-Aware Curriculum Learning method (HACL) as a meta-scheduler that incorporates the reward history\eqref{eq:hidden} via a recurrent model. HACL acts as a meta-policy ($\pi_{\text{meta}}$), which selects curriculum bins to guide the low-level control policy ($\pi_{\theta}$). Using unmodeled history, HACL improves gait stability and speed at higher command velocities.

\subsection{Problem Set-up (Tasks \& Design Space)}
Rather than injecting the history ht into the main control policy ($\pi_{\theta}$), which will make it computationally expensive and difficult to understand the impact of history, we isolate the history at the curriculum level. At the start of each rollout (episode), a meta-policy selects bins $b_t$  \eqref{eq:meta_policy} from a discrete set $\mathcal{B}$ using the summarized reward history: 

\begin{equation}
b_t \sim \pi_{\text{meta}}(\,\cdot \mid h_{t-1}), \quad b_t \in \mathcal{B}
\label{eq:meta_policy}
\end{equation}

Thus, $\pi_{meta}$ schedules episode-level bins (this is not MetaRL \cite{vanschoren2019meta}). By decoupling meta-scheduling from the main low-level controller, which remains a feed forward actor-critic, we can isolate the benefits of reward history in quadrupedal locomotion. The low-level policy ($\pi_{\theta}$) receives input from proprioceptive observation ($o_t$) (for example, joint position $({q}_t)$ and joint velocity $({\dot{q}}_t)$) and outputs the parameterized joint position commands ($q^{des}_t$) (Figure 2). To analyze $\pi_{\text{meta}}$, we consider velocity tracking with commanded velocity tasks $(v_{x}^{\text{cmd}}, v_{y}^{\text{cmd}}, \omega_{z}^{\text{cmd}})$. In Isaac Gym \cite{makoviychuk2021isaac}, we discretize this 3D command space into a set of bins ($(b_t)$), the union of which is represented by $\mathcal{B}$ \eqref{eq:bins}.

\begin{equation}
    B = \{b_1, b_2, \dots, b_N\}.
\label{eq:bins}
\end{equation}

with N equal to 4000 (details in the experiment section). From these bins, the agent samples and executes target command velocities (linear ($v^{cmd}_x$), and angular ($\omega^{cmd}_z$)).

Previous works \cite{margolis2024rapid}, \cite{kumar2021rma}, \cite{hwangbo2019learning} find training succeeds when commands are selected from a small uniform probability distribution range of $[-1.0, 1.0]$ but degrade or fail for a large range (Fig. 3, Section IV.A). Consequently, HACL also begins with low ranges and gradually expands. The meta-policy $\pi_{\text{meta}}$ scores the bins using the reward history and chooses the bins with the highest predicted return while ignoring the low-return bins. This history-aware selection handles temporal drift and partial observability \cite{zhao2020non} better than the random or fixed schedule curricula, improving speed and stability as the range expands (Table II).

\begin{table}[t]
\centering
\begin{tabular}{@{}lp{0.65\columnwidth}@{}} % Adjust column width
\toprule
\textbf{Symbol(s)} & \textbf{Description} \\ \midrule
$v_x$ ,$v_y$ $\omega_z$  & Observed linear (x,y) and angular velocities respectively.  \\ 
$v^{cmd}_x$ ,$v^{cmd}_y$ $\omega^{cmd}_z$   &  Linear (x,y) and angular command velocity of the robot base. \\ 
$\hat{r}_{\text{lin}}$ , $\hat{r}_{\text{ang}}$ & RNN predicted linear velocity and angular velocity rewards respectively. \\
${r}_{\text{lin}}$ , ${r}_{\text{ang}}$, $r_t$ & Observed linear and angular velocity rewards, and rewards at time t. \\
% $\tau$ & Torque for the each robotic joints. \\
$\tau$ , g \& $g_{\psi}(\cdot)$ & Torque, gravitational acceleration \& predictor head \\
${b_1, b_2, \dots, b_n}$ & The discretized design space bins totaling 4000 bins\\
% ${w_1, w_2, \dots, w_n}$ & The normalized probability weights distribution for bins\\
$\hat{\mu}(b)$ & The predictions made by RNN. \\
% $\hat{\mu}(b)$, $\mu_t(b_t^*)$ & The predictions made by RNN and the best possible bin during a timestep for rewards\\
$h_{t-1}$ & The RNN hidden state for the time-step $t-1$.\\

$x_t$, $c_t$ & One hot encoded vector and sampled command.\\
$w_t (b)$ \& $\pi_{\text{meta}}(b)$  & Weights and normalized distribution.\\
$f_{\text{RNN}}(h_{t-1}, [x_t, r_t]) $ & The RNN function with hidden layer ($h_{t-1}$) and $x_t$, $r_t$ as inputs.\\

% $R(T)$ & The cumulative regret over time step $T$.\\

$L(\psi)$ & The loss function with variable $\psi$. \\

$q_t$ , $\dot{q}_t$ , $q^{des}_t$& The robot's joint position, joint velocity and task parameter joint command at time $t$.\\

CoT, $S$ & Cost of Transport and the stability score.\\

% $DOF$ & Robot's degrees of freedom.\\

\bottomrule
\end{tabular}
\caption{List of key symbols and their descriptions}
\label{tab:key symbols}
\end{table}

\subsection{Partially Observed Nature of Legged Locomotion} 

Legged robots such as Go1 sense joint states, IMU data, and contact forces, but friction, sensor latency, and actuator dynamics are unobservable, making locomotion a partially observed Markov decision process (POMDP). Previous works \cite{zhao2020non}, \cite{yu2019sim}, \cite{byl2008metastable} show that modeling history is crucial for rapid and agile maneuvers; unmodeled latent factors can affect a robot's stability, speed, and energy efficiency (Table II). During training, we capture non-Markovian effects by modeling the reward history with a recurrent neural network \cite{rumelhart1985learning}, updated as \eqref{eq:hidden}:

% \begin{equation}
% Q^{\pi}(s_t,a_t \mid h_{t-1})
%   = r_t+\gamma\,\mathbb E_{a'\!\sim\!\pi}
%     \bigl[\,Q^{\pi}(s_{t+1},a' \mid h_t)\bigr]
% \tag{2}
% \end{equation}

\begin{equation}
h_t = f_{\text{RNN}}(h_{t-1}, [x_t, r_t])
\label{eq:hidden}
\end{equation}

The hidden state $(h_{t-1})$ summarizes the previous bins and the respective rewards during each episode, allowing the meta-policy ($\pi_{meta}$) to help account for the long-term effects of velocity commands based on the observed rewards (${r}_{lin}$, ${r}_{ang}$): effects that a fixed update or other similar curriculum methodology typically miss \cite{margolis2024rapid}, \cite{rudin2022advanced}, \cite{kumar2021rma}. RNNs (e.g., LSTM/GRU/Vanilla RNN) are well suited to such dependencies because the hidden state $(h_{t-1})$ compacts past information. We encode the bins with $\text{BinID}_t$ as one hot vector $x_t$ \eqref{eq:one-hot},

\begin{equation}
x_t = \text{one-hot}(\text{BinID}_t) \in \{0,1\}^N,\qquad N = 4000
\label{eq:one-hot}
\end{equation}

And predict the expected reward $\hat{\mu}(b_t)$ for any bin $b_t$ via \eqref{eq:predictions},
\begin{equation}
\hat{\mu}(b_t) = g_\psi(h_{t-1}, x_t) = \big[\, \hat{r}_{\text{lin}}(b), \; \hat{r}_{\text{ang}}(b) \,\big]
\label{eq:predictions}
\end{equation}

Here $\hat{r}_{\text{lin}}(b)$ and $\hat{r}_{\text{ang}}(b)$ are the predicted linear and angular rewards for the bins, respectively (bin sizes and command ranges are discussed in detail in Section IV).

% \begin{equation}
% h_t = \text{LSTM}(h_{t-1}, [x_t, r_t(b_t)], \theta).
% \label{eq:LSTM}
% \end{equation}

\subsection{History-Aware Sampling}

Because bins are scored with the reward history, the sampler must avoid getting stuck in low rewarding regions while also not blindly discard bins that may yield higher return later. We map the reward predictions $\hat\mu_b$ from Equation \eqref{eq:predictions} to a scalar utility function defined by \eqref{scalar}

\begin{equation}
u(b) = \alpha \,\hat{r}_{\text{lin}}(b) + (1-\alpha)\,\hat{r}_{\text{ang}}(b),
\quad \alpha \in [0,1]\;
\label{scalar}
\end{equation}

and update bins via clipped History-Aware greedy rule \eqref{eq:ha greedy}:

\begin{equation}
   w_t(b) = \operatorname{clip}\!\big(w_{t-1}(b) + \kappa\, u(b), \, 0, \, 1\big),
\quad \kappa = 0.2
\label{eq:ha greedy}
\end{equation}

\begin{figure}[thpb]
  \centering
  \includegraphics[width=0.9\columnwidth]{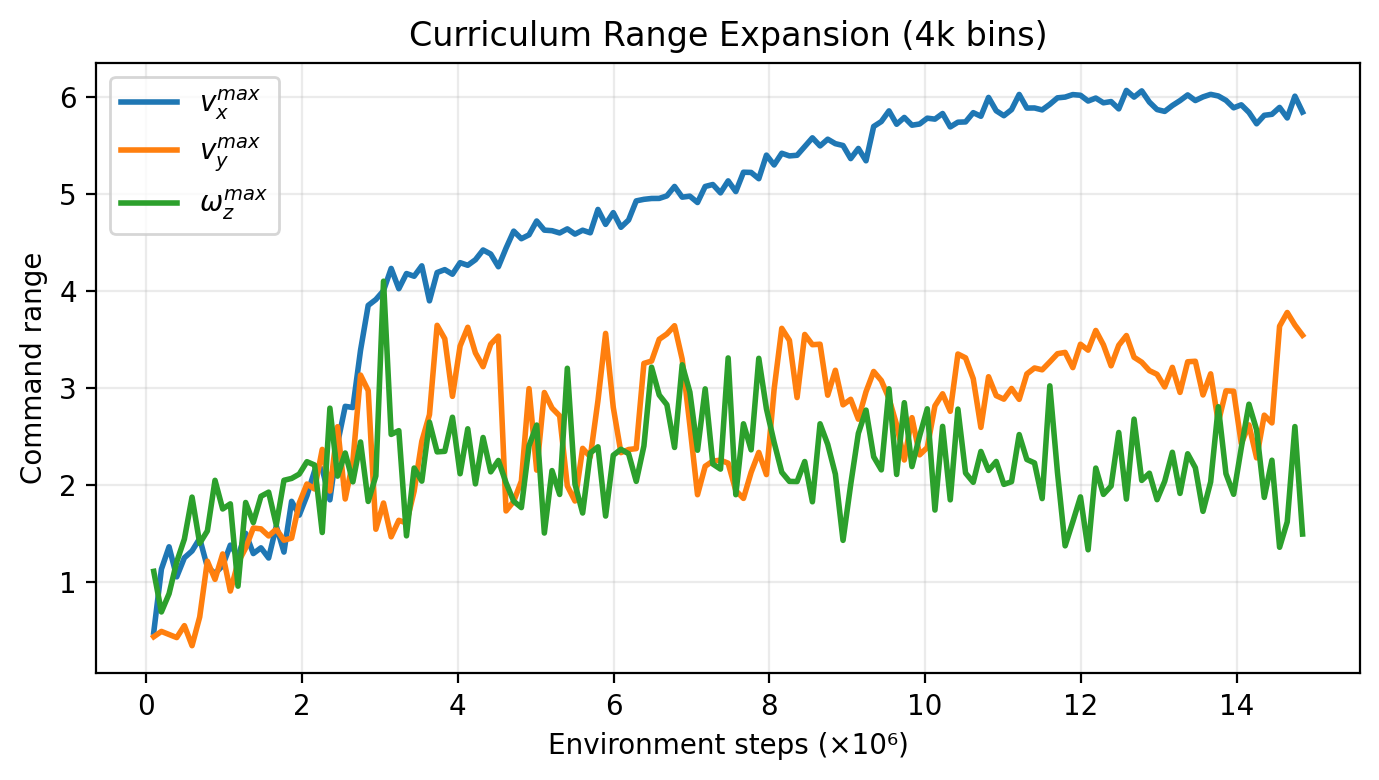}
  \caption{\textbf{Curriculum Range Expansion:} 
  HACL leverages history to achieve better results by allowing the agent to master the lower range first before proceeding to higher velocities, avoiding pitfalls of instability and failure in the training process. Curriculum begins with the range $[-1.0,1.0]$  and expands by  $[-0.5, 0.5]$. 
  As evident from the plot, the policy learns \textcolor{blue}{$v_x^{\text{max}}$} faster and more steadily, while \textcolor{orange}{$v_y^{\text{max}}$} and \textcolor{green}{$\omega_z^{\text{max}}$} are less consistent and saturate at lower thresholds, owing to our reward optimization of forward locomotion.}
  % eventually all bounds saturate at the defined task maxima and covering the defined command space.
  \label{fig:curr_rang_overview}
\end{figure}

Here, $\kappa > 0$ is the step size. Using bin weights with predicted return, HACL exploits high-value regions while maintaining exploration, improving long-horizon return (Table II). Intuitively, the bins with higher predicted rewards are selected more often, so their commands are sampled more frequently. Because bins encode command velocities, these shifts can be seen in terms of improving velocity and stability. We implement this at the meta-level ($\pi_{meta}$) by normalizing the weights per bin \eqref{eq:command_sampling}

\begin{equation}
\pi_{\text{meta}}(b) = 
\frac{w_t(b) + \varepsilon}{\sum_{b' \in B} \big(w_t(b') + \varepsilon\big)},
\quad \varepsilon =10^{-3}
\label{eq:command_sampling}
\end{equation}

Equation \eqref{eq:command_sampling2} implements two-stage sampling: first draw a bin ($b_t$) from
$\pi_{meta}$; then sample a command value within the active command range.

\begin{equation}
b_t \sim \pi_{\text{meta}}(\,\cdot\,), 
\qquad 
c_t \sim \operatorname{Uniform}\!\big(b_t \cap V\big)
\label{eq:command_sampling2}
\end{equation}

% \begin{equation}
%     w_{t+1}(b) = w_{t}(b) + \alpha \left(\hat{r}_{\text{lin}}(b) + \hat{r}_{\text{ang}}(b)\right),
% \label{eq:weight update}
% \end{equation}

with V being the upper range for those command values. As training progresses, observed rewards update the predictor, sharpening $\hat{\mu}(b_t)$, increasing the weight for high-value bins, and accelerating convergence, leading to higher rewards with better stability.

% \begin{equation}
% \mathbf{v}^{\max}_{k+1} =
% \begin{cases}
% \min\bigl(\mathbf{v}^{\max}_{k} + \boldsymbol{\Delta},\; \mathbf{V}\bigr),
%    & \text{if success},\\[4pt]
% \mathbf{v}^{\max}_{k}, & \text{otherwise},
% \end{cases}
% \label{eq:update}
% \end{equation}
% %

% where $\mathbf{v}^{\max}_{k} = [v_{x,k}^{\max},\, v_{y,k}^{\max},\, \omega_{z,k}^{\max}]^{\top}$, $\boldsymbol{\Delta} = [\Delta v,\, \Delta v,\, \Delta \omega]^{\top}$, and $\mathbf{V} = [V_x,\, V_y,\, \Omega]^{\top}$.

\begin{equation}
\begingroup
\setlength\abovedisplayskip{3pt}
\setlength\belowdisplayskip{3pt}
\small
\mathbf{v}^{\max}_{k+1} =
\begin{cases}
\min\!\bigl[\mathbf{v}^{\max}_k + (\Delta v,\Delta v,\Delta\omega),\; \mathbf{V}\bigr], & \text{if success} ,\\
\mathbf{v}^{\max}_k, & \text{otherwise.}
\end{cases}
\label{eq:expansion}
\endgroup
\end{equation}

where $V$  and $\mathbf{v}$ are the upper and lower limit ranges of command velocities, respectively, and we define success when rewards exceed the defined threshold (Eq. \eqref{eq:expansion} and Fig. 3).

% ($(V_x, V_y, \omega_{z})$ and $(v_x^{\max}, v_y^{\max} \omega_z^{\max})$), 

\begin{figure}[h]
    \centering

     % Second row of images
    \begin{minipage}{0.42\linewidth}
        \centering
        \includegraphics[width=\linewidth]{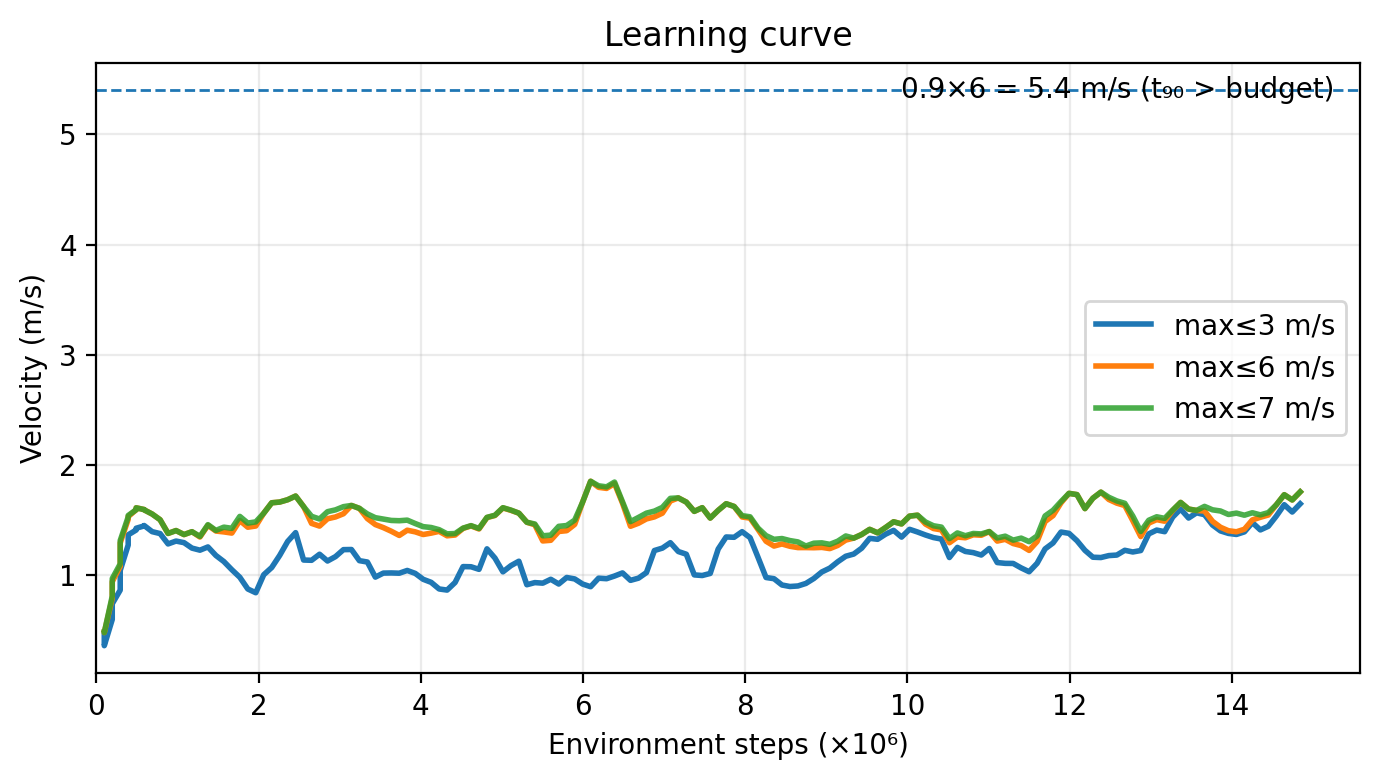}
    \end{minipage}
    \begin{minipage}{0.42\linewidth}
        \centering
        \includegraphics[width=\linewidth]{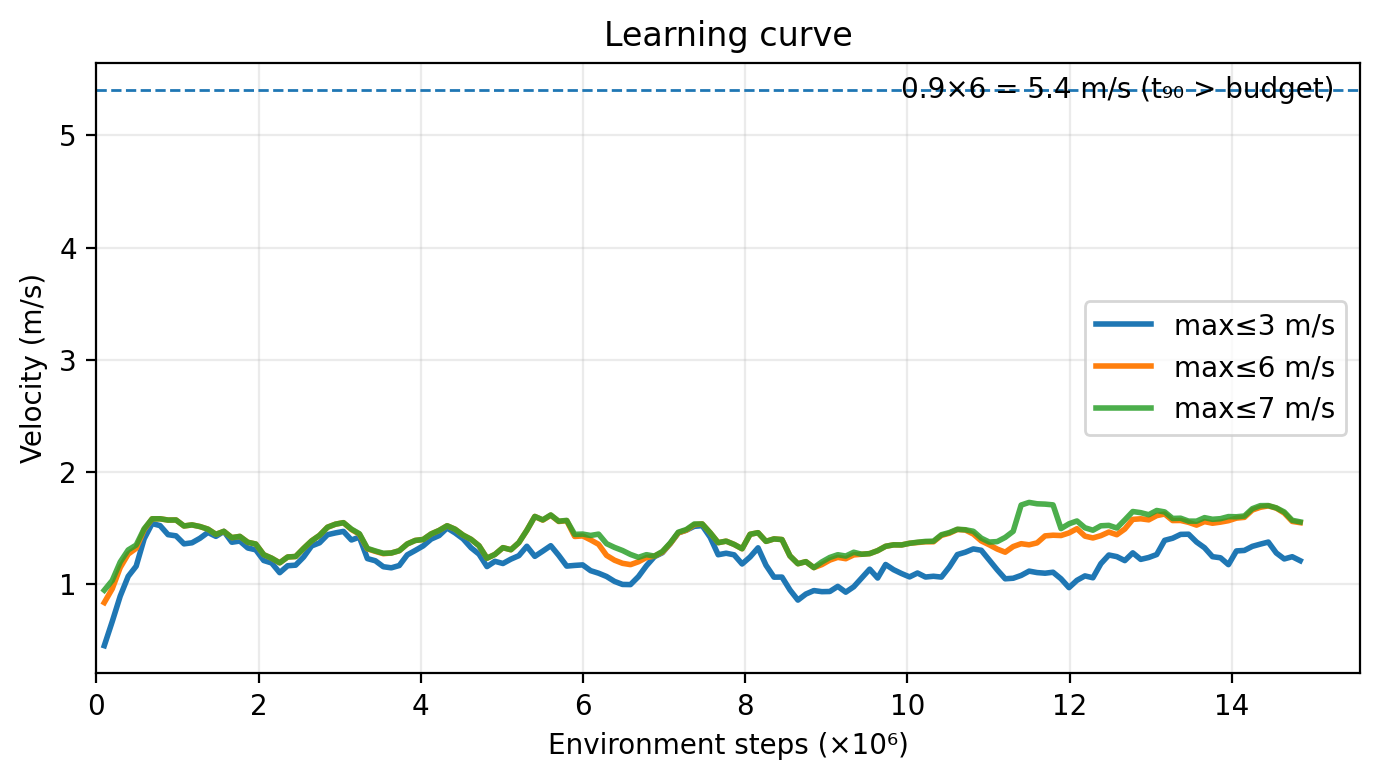}
    \end{minipage}
  
      % Second row of images
    \begin{minipage}{0.42\linewidth}
        \centering
        \includegraphics[width=\linewidth]{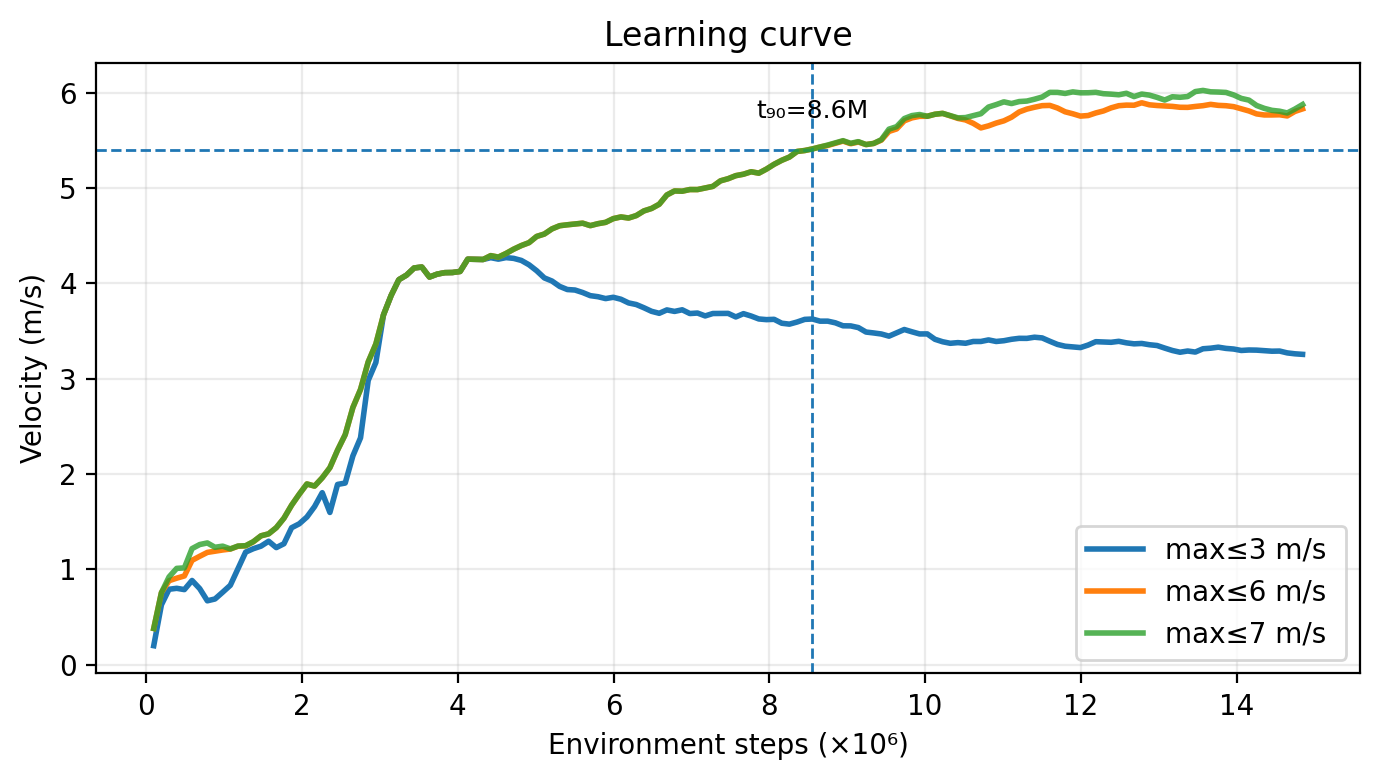}
    \end{minipage}
    \begin{minipage}{0.42\linewidth}
        \centering
        \includegraphics[width=\linewidth]{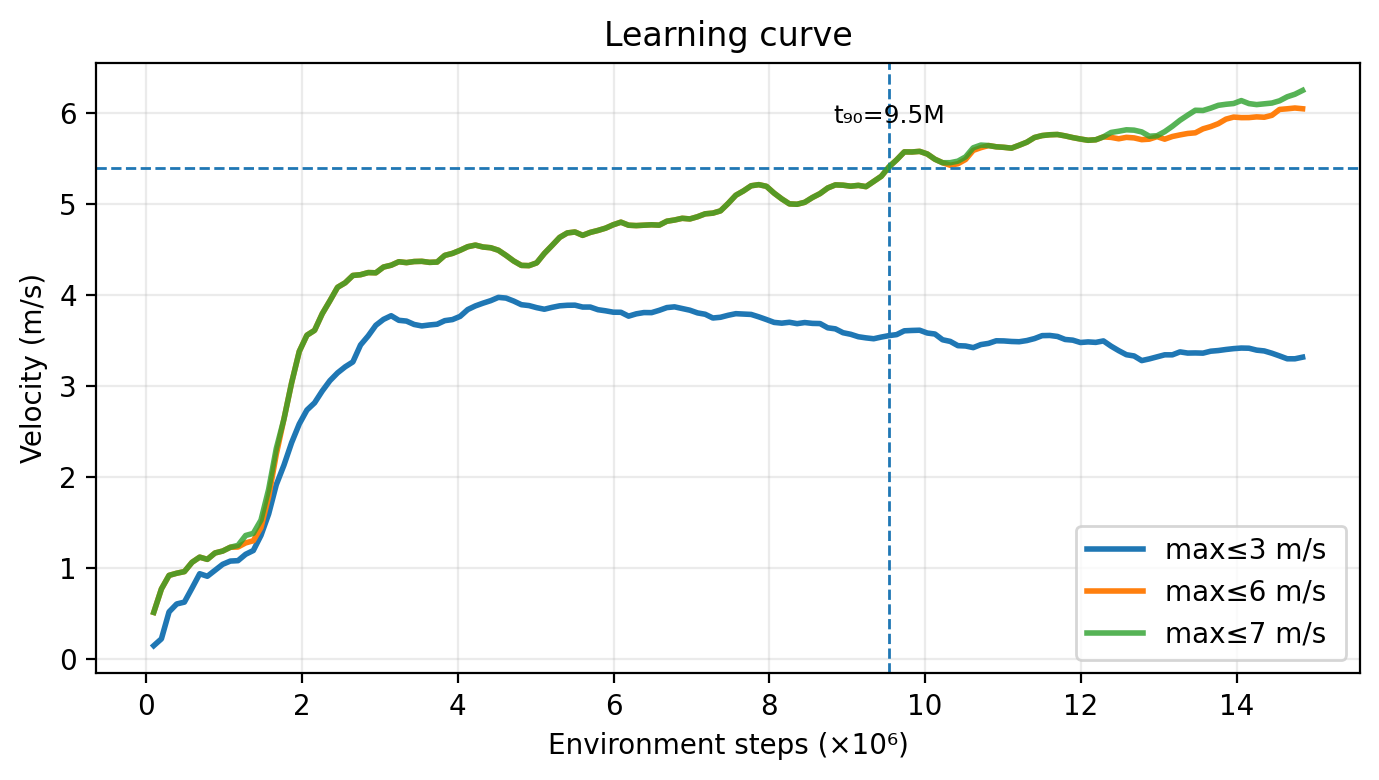}
    \end{minipage}           
   
    \caption{\textbf{Binning Criteria:} HACL trained for 250, 1000, 4000 and 6000 bins with identical training conditions to measure efficiency and performance in 1500 iterations. Coarse binning like 250 bins and 1000 bins fails to learn high-speed locomotion and gets stuck between the range of 1.5 to 2 m/s. With 4000 bins the robot achieves 6 m/s and reaches 90\% of target velocity within 8.6 million steps whereas with 6000 bins robot reaches 6-6.3 m/s and the 90\% of target velocity within 9.5 million steps, basically needs more training time to visit each bin leading to greater reward variance.}
    \label{fig:hardware_robust}
\end{figure}

\subsection{Training Objective}

To model non-Markovian reward dynamics, we use a recurrent predictor, which retains information over time through a hidden state $h_{t-1}$ and captures temporal history in curriculum design, and is trained to minimize \eqref{eq:loss}:

\begin{equation}
\begin{aligned}
L(\psi) \;=\; \frac{1}{T} \sum_{t=1}^{T} 
   \big\|\, r_t - \hat{\mu}(b_t) \,\big\|^2
\label{eq:loss}
\end{aligned}
\end{equation}

prediction error between the observed rewards $(r_t)$ and the predicted RNN rewards $\hat{\mu}(b_t)$. As the RNN collects more data, its prediction of the bins improves, enabling more efficient bin selection and faster convergence to higher forward velocities. Minimizing $L(\psi)$ improves $\hat{\mu}(b_t)$, which in turn drives $\pi_{meta}$ towards a low regret bin schedule that improves the speed and stability of the agent.

\begin{figure*}[!t]
    \centering

    % ---- Top row: action frames ----
    \begin{minipage}{\textwidth}
        \centering
        \includegraphics[width=0.19\textwidth]{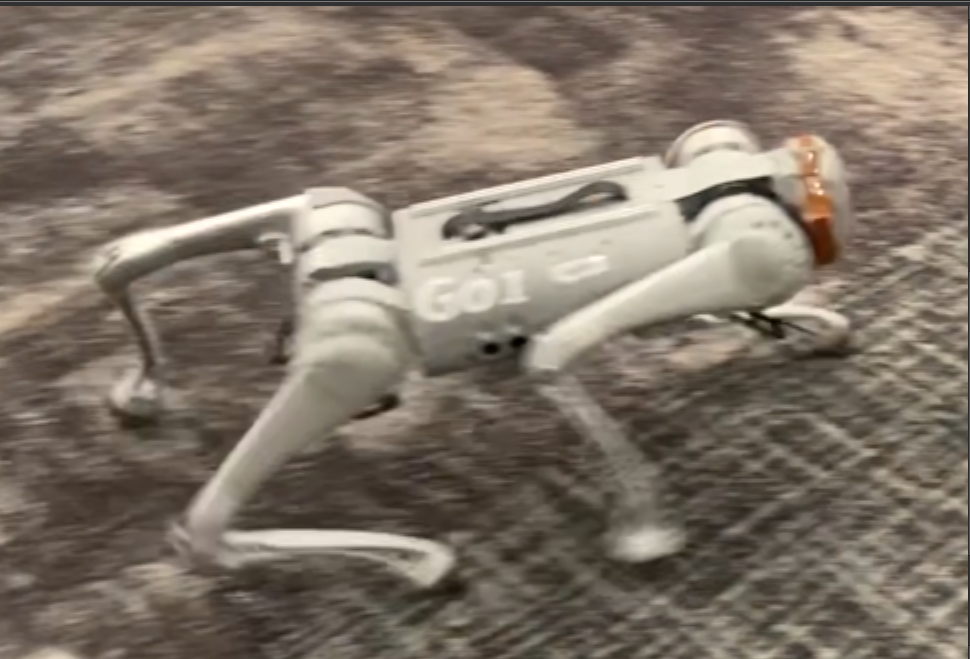}\hfill
        \includegraphics[width=0.19\textwidth]{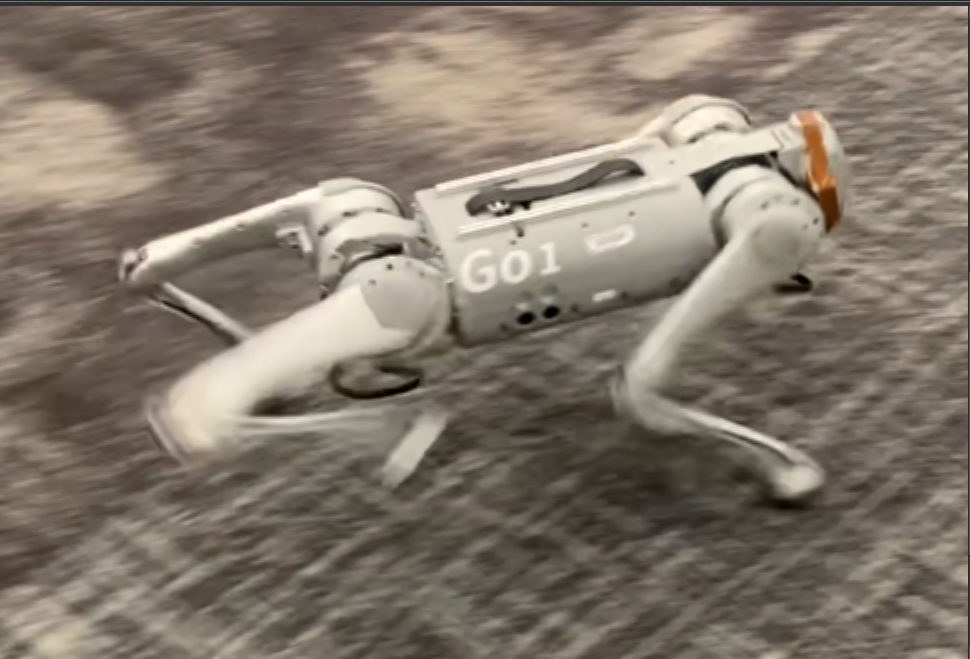}\hfill
        \includegraphics[width=0.19\textwidth]{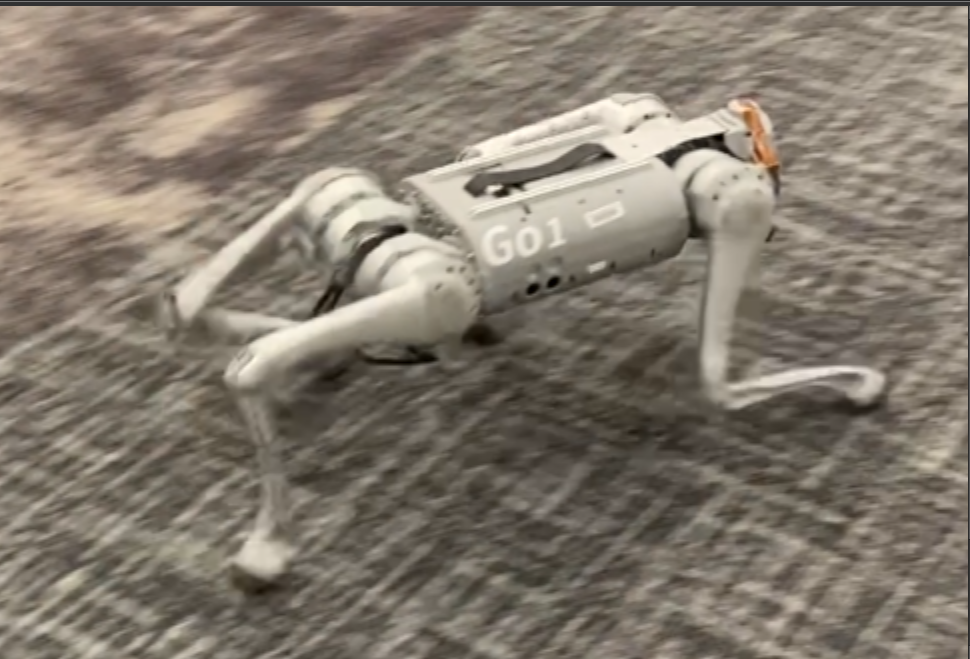}\hfill
        \includegraphics[width=0.19\textwidth]{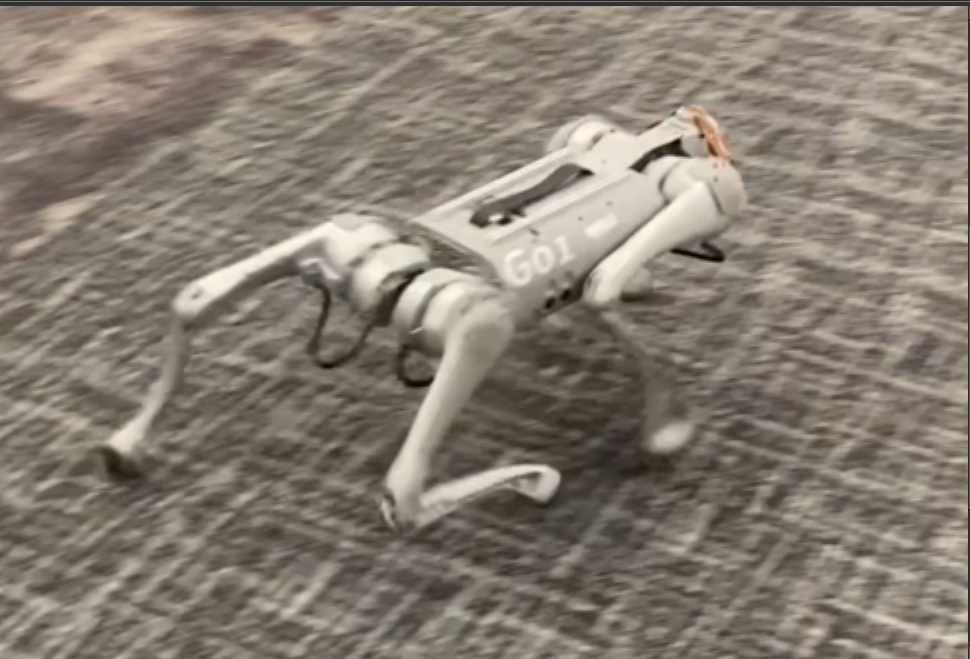}\hfill
        \includegraphics[width=0.19\textwidth]{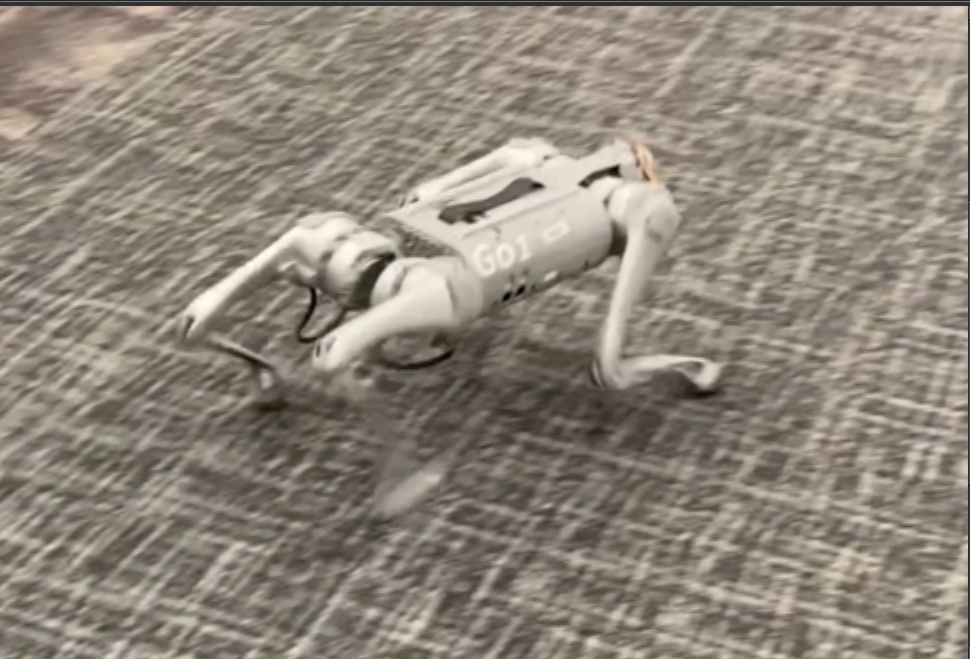}
    \end{minipage}

    % ---- Velocity plot ----
    \vspace{-0.05em} % reduce vertical gap
    \begin{minipage}{\textwidth}
        \centering
        \includegraphics[width=1.00\textwidth]{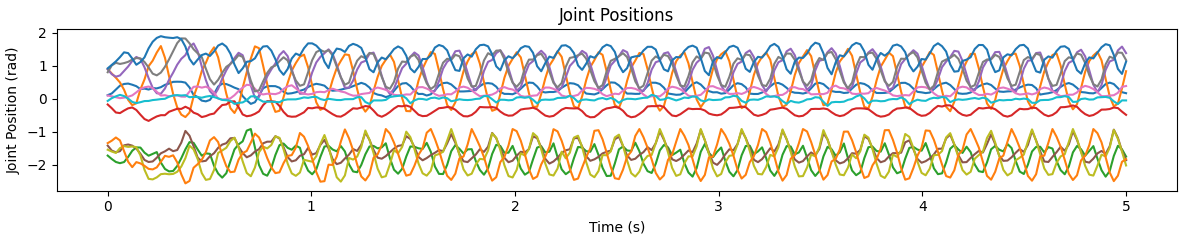}
    \end{minipage}

    % % ---- Joint plot ----
    % \vspace{-0.05em}
    % \begin{minipage}{\textwidth}
    %     \centering
    %     \includegraphics[width=1.00\textwidth]{img/gabeJoint7ms.png}
    % \end{minipage}

    \caption{\textbf{High-speed locomotion with HACL.} 
    \textbf{Row 1 (Real-world, HACL):} On woolen carpet policy achieves $4.1 \pm 0.2$ m/s over a 3-8 m run with a task success rate of 90\% .
    \textbf{Row 2 (Sim, HACL):} On command velocity of 7 m/s, Go1 reaches $6.7 \pm 0.2$ m/s and the joint-position time plot shows a stable leg coordination(12 joints: LF, RF, LH, RH )  at higher command velocity. HACL maintains stable, consistent joint position both in the real-world and in simulation (\textbf{Row 1 \& 2})} 
    \label{fig:knit_fullwidth}
\end{figure*}

\section{Experiments}

\subsection{Simulation Environment}
\textbf{Simulation Details.} We build our code in open-source repositories \cite{margolis2024rapid}, \cite{rudin2022learning} and model Unitree Go1 with a total of 12 Degrees of Freedom (DoF) using the URDF in the IsaacGym simulator \cite{makoviychuk2021isaac}. We then added Unitree Go2 and MIT Mini Cheetah (each trained independently to test generalization). Training uses $4000$ parallel environments at a time step of \(\mathbf{\Delta t}=5\;{\rm ms}\) on an Nvidia RTX 4090 laptop-based GPU. Experiments are run over 400 million steps and 4000 PPO updates (roughly ~4 hours).

\textbf{Command Curriculum.} We sampled linear and angular command targets $ v^{cmd}_x$ and $ \omega^{cmd}_z$ from the lower range $[-1.0,1.0]$. Similarly to \cite{margolis2024rapid}, we have also observed that a wide initial range, such as $[-5,5]$, destabilizes learning and prevents the robot from reaching even moderate speed ($\approx 4 \,\text{m/s}$). In contrast, starting with lower ranges and gradually expanding by $[-0.5,0.5]$ produces a more stable and higher speed, as shown in Figure 3. 

\textbf{Discretization.} We discretize the task parameter space of the IsaacGym simulator into a 3D grid. The X, Y, and Z axes are normalized to $[-1, 1]$ and divided into $20 \times 10 \times 20$ bins, generating a total of 4000 bins. We have empirically found that the use of 4000 bins is the sweet spot for sample efficiency, speed, stability, and outperforms both coarser (250, 1000) and finer (6000) (Figure 4).

\textbf{Domain Randomization (DR).} To improve sim-to-real transfer, we randomize key dynamics and sensor parameters (e.g., joint friction, motor delays, and sensor noise)\cite{tobin2017domain}. In our DR ablation, we first tested randomization in simulation on a fixed rule-based curriculum \cite{margolis2024rapid}, \cite{rudin2022learning} by expanding domain parameter ranges and without using a reward history / HACL. In this setting, the policy achieves a speed of 5.3 m/s slower than the standard 5.45 m/s \cite{margolis2024rapid} for a command velocity of 6 m/s. We hypothesize that over-randomization of parameters leads to a conservative policy and reduces the speed, consistent with \cite{margolis2024rapid} \cite{tan2018sim}. We therefore revert to the original parameter ranges \cite{margolis2024rapid} with minor tweaks. Unless explicitly mentioned, most of our domain parameters and reward functions follow \cite{margolis2024rapid} \cite{rudin2022learning}, with minor tweaking, and remain the same across all baseline comparisons.

% \cite{xie2021dynamics}

\subsection{Experimental Setup (Teacher-Student Policy)}

Many real-world parameters, such as terrain type, slope, friction coefficient, etc., are often unknown and unobservable. To overcome this, we have utilized teacher-student training. Following \cite{margolis2024rapid}, \cite{rudin2022learning}, we train a teacher policy ($\pi_{T}$) which takes the privilege parameters encoded by $e_{\theta}$ (for example, friction, restitution, base mass, joint friction, etc.) along with current observation $o_t$. The embeddings are passed into a multilayer perceptron of size $[512,256,128]$. Because these parameters are not available in the real world, we train the student policy ($\pi_{s}$) to infer them through a history of the last $h$ timestep $H = [x_{t-h}, x_{t-1}]$ for system identification \cite{kumar2021rma}, \cite{lee2020learning} and mimic the teacher ($\pi_{T}$) during deployment. The state history ($H$), which is used for System-ID is different from the reward history used by the curriculum to reweight command sampling (History-Aware Greedy (HA-Greedy)). The low-level actor critic policy is optimized with PPO \cite{schulman2017proximal}. HACL acts as meta-level sampler for curriculum, and the gradients from the low-level policy do not back-propagate to the HACL module. We did not feed the reward history to either $\pi_{T}$ or $\pi_{S}$,  as that would muddle the real benefits of the reward history. So, by isolating it at the curriculum level, we can better understand the curriculum; learning process; and its utility for bin selection. (ref. Table I for symbols and description).

\begin{table*}[t]
\centering
\caption{HACL vs. curriculum baselines. Simulation results of replicated SOTA curriculum methods (\,\textsuperscript{‡}\,) ; Zero-shot transfer and max. real world speed (if reported) (\,\textsuperscript{†}\,).}
\label{tab:method_comparison}
\renewcommand{\arraystretch}{1.1}
\setlength{\tabcolsep}{4pt}
\begin{tabular}{lcccccccc}
\toprule
Method &
Adaptation &
\begin{tabular}[c]{@{}c@{}}Curric.\\History?\end{tabular} &
\begin{tabular}[c]{@{}c@{}}Range\\Exp.\end{tabular} &
\begin{tabular}[c]{@{}c@{}}Curric.\\Decoupled?\end{tabular} &
\begin{tabular}[c]{@{}c@{}}Sim $V^{cmd}_x$ \textsuperscript{‡}\\ $=7$ (m/s)\end{tabular} &
\begin{tabular}[c]{@{}c@{}}Stability \textsuperscript{‡} \\$S$\end{tabular} &
\begin{tabular}[c]{@{}c@{}}Zero-shot\textsuperscript{†}\\robot\end{tabular} &
\begin{tabular}[c]{@{}c@{}}Max real \textsuperscript{†} \\speed (m/s)\end{tabular} \\
\midrule
RMA~\cite{kumar2021rma} &
Inputs/Params &
% \textsc{None} &
None &
No &
No &
4.7 & 1200 & A1 & — \\
RLvRL~\cite{margolis2024rapid} &
Velocity commands &
% \textsc{Obs-Reward} &
Obs-Reward &
Yes &
% \textsc{Decoupled} &
Yes &
5.7 & 1400 & Mini Cheetah & 3.9 \\
Aractingi~\cite{aractingi2023controlling} &
Tasks/Terrain &
% \textsc{None} &
None &
Hybrid &
% \textsc{Coupled} &
No &
5.5 & 1100 & Solo12 & 1.5 \\
Hutter ~\cite{zhang2024learning} &
Tasks/Terrain &
% \textsc{Obs--Reward} &
Obs-Reward &
% \textsc{None} &
No &
% \textsc{Coupled} &
No &
— & — & ANYmal & $\ge 2.5$ \\
ALP--GMM~\cite{portelas2020teacher} &
Tasks/Commands &
% \textsc{Obs--LP} &
Obs-LP &
% \textsc{Implicit} &
No &
% \textsc{Decoupled} &
Yes &
3.5 & 750 & No & — \\
Self-Paced~\cite{kasaei2020learning} &
Tasks/Commands &
% \textsc{Obs--Reward} &
Obs-Reward &
% \textsc{Implicit} &
No &
% \textsc{Coupled} &
Yes &
2.6 & 700 & No & — \\
UCB~\cite{auer2002using} &
Tasks/Commands &
% \textsc{Obs--Reward} &
Obs-Reward &
% \textsc{Utility} &
Yes &
% \textsc{Decoupled} &
Yes &
0.2 & 780 & No & — \\
Thompson~\cite{chapelle2011empirical} &
Tasks/Commands &
% \textsc{Obs--Reward} &
Obs-Reward &
% \textsc{Utility} &
Yes &
% \textsc{Decoupled} &
Yes &
0.11 & 640 & No & — \\
% \rowcolor{black!5}
\textbf{HACL (Ours)} &
Meta-sampler &
% \textsc{Pred--RNN} &
Pred-Reward &
% \textsc{Utility} &
Yes &
% \textsc{Decoupled} &
Yes &
\textbf{6.7} & \textbf{2000} & Go1 & \textbf{4.1} \\
\bottomrule
\end{tabular}

\vspace{2pt}
\footnotesize
% \textsuperscript{†}\,From the cited paper. \;
\textsuperscript{‡}\,Our implementation of SOTA methods in the HACL pipeline on Go1 (Isaac Gym; $n{=}5$. \;
$S$ = \emph{Stability score, higher is better (Ref. C. Eval. Metric)}.\\
\emph{Legend:} None = fixed curriculum schedule (no history); 
Obs-Reward= observed reward; 
Obs-LP = observed learning progress; 
Pred-Reward = RNN-predicted rewards; 
% \textsc{Thresh} = neighbor add when reward $\ge\gamma$; 
% \textsc{Implicit} = distribution drifts outward (no explicit bounds);
% \textsc{Utility} = expansion by top-utility/frontier bins; 
Decoupled = separate sampler with no shared weights with low-level policy.
\end{table*}

\subsection{Evaluation Metric}

\textbf{Cost of Transport (CoT)} \cite{schperberg2025energy}. We define CoT for the Go1 robot as the energy or power consumed by each joint at each timestep \eqref{eq:cot}: 

% One of the key metrics for our evaluation is the Cost of Transport (CoT) \cite{schperberg2025energy} of the Go1 robot. The energy or power consumed by each joint at each timestep is given by the torque required for each joint at the timestep, i.e. $\tau_j(i)$ multiplied by $\dot{q}_j(i)$, the joint velocity for each joint at timestep $i$. The power is divided by the weight multiplied by the distance traveled per time step and is given by $mg \, \Delta s$ using \eqref{eq:cot}

\begin{equation}
\mathrm{CoT} = \frac{\int_0^T \sum_{j=1}^{N} \tau_j(t) \, \dot{q}_j(t) \, dt}{mg \, \Delta s}.
\label{eq:cot}
\end{equation}

Where $\tau_j(i)$ and $\dot{q}_j(i)$ are the torque and velocity of the joint for each joint at the time step $i$, and $mg \, \Delta s$ is the distance traveled per time step. 

\textbf{Stability score (S).} We define it as \eqref{eq:stability}:

\begin{equation}
S \;=\; 
\frac{1}{N}
\sum_{i=1}^{N} 
\sum_{k=1}^{K} 
W_k\, r_k(i),
\label{eq:stability}
\end{equation}

Where the rewards $r_k(i)$ measure orientation, base height, angular velocity, linear velocity, joint position limit, joint velocity limit, raw joint velocity, self-collision and torque limit constraints at any time step, and $W_k$ are fixed weights. Higher $S$ indicates higher stability.

% \begin{equation}
% \begin{aligned}
% S = \sum_{i=1}^{N} \Big( 
%     & r_{\text{orient}}(i) + r_{\text{base height}}(i) + r_{\text{ang vel xy}}(i) \nonumber \\
%     & + r_{\text{lin vel z}}(i) + r_{\text{dof pos limits}}(i) + r_{\text{dof vel limits}}(i) \nonumber \\
%     & + r_{\text{dof vel}}(i) + r_{\text{collision}}(i) + r_{\text{torque limits}}(i) 
%     \Big)
% \end{aligned}
% \label{eq:stability}
% \end{equation}

\textbf{Task success rate.} It is the percentage of successful runs without a crash, tip, faulty leg coordination or fall. We report the mean of 10 trials in the real world and 5 in simulation on the Go1 robot, with 95\% confidence interval.

% \begin{equation}
% \Delta R_t(b) = G_t(b) - \bar{G}_t
% \label{regret}
% \end{equation}

% And we have analyzed $\Delta R_t(b)$ based on the final rewards received by HA-Greedy, Thompson sampling and UCB etc.

\subsection{Baselines}

\textbf{Curriculum baselines:} We have compared our HACL method against the following baselines: (i) Rapid-Motor Adaptation (RMA) \cite{kumar2021rma}: fixed schedule curriculum that gradually increases penalties (ii) Grid-Adaptive curriculum  \cite{margolis2024rapid}: fixed rule-based bin selection (iii) Solo12 \cite{aractingi2023controlling}, risky terrains\cite{zhang2024learning}: terrain curriculum (iv) Gaussian Mixture models (ALP-GMM) \cite{portelas2020teacher}:  adaptive curriculum (v) While Self-paced learning \cite{kasaei2020learning}: paces tasks based on value function. None of these approaches considers history.

\textbf{Bandits. } Thompson Sampling (TS) \cite{chapelle2011empirical} balances exploration and exploitation, but in our case increases the training time (550 minutes) and converges slowly, probably due to overexploration and hyperparameter sensitivity without improving the outcome. TS maintains per bin count and equation \eqref{eq:ha greedy} becomes \eqref{eq:weight ts}:

\begin{equation}
\theta_b \sim \operatorname{Beta}(\alpha_b,\beta_b),\;
   w_t(b) = \theta_b
\label{eq:weight ts}
\end{equation}

\textbf{UCB }(Upper Confidence Bound) assumes stationarity of the rewards and might over-penalize bins that might become valuable as gait improves.  The bin weights in case of UCB are given by \eqref{eq:weights ucb}:

\begin{equation}
w_{t}(b) = \hat{\mu}_{b} + \sqrt{\frac{2 \ln t}{n_{b}}}.
\label{eq:weights ucb}
\end{equation}

where $\hat{\mu}_{b}$ and $n_{b}$ are the empirical mean and the selection count, respectively. 

\textbf{Non-history neural nets.} We also include CNN and MLP feedforward schedulers.  In our findings, they outperform UCB and Thompson Sampling, despite ignoring history.  The likely reason is better gradient optimization on reward signals and that they extract better features (Tables II and III). 

% \begin{table}[h]
% \caption{An Example of a Table}
% \label{table_example}
% \begin{center}
% \begin{tabular}{|c||c|}
% \hline
% One & Two\\
% \hline
% Three & Four\\
% \hline
% \end{tabular}
% \end{center}
% \end{table}

%    \begin{figure}[thpb]
%       \centering
%       \framebox{\parbox{3in}{We suggest that you use a text box to insert a graphic (which is ideally a 300 dpi TIFF or EPS file, with all fonts embedded) because, in an document, this method is somewhat more stable than directly inserting a picture.
% }}
%       %\includegraphics[scale=1.0]{figurefile}
%       \caption{Inductance of oscillation winding on amorphous
%        magnetic core versus DC bias magnetic field}
%       \label{figurelabel}
%    \end{figure}

\section{Results \& Discussion}

\subsection{Simulation Learning}

For a command velocity of 7 m/s, HACL reaches 6.7 m/s and outperforms all methods (Table II).  Although prior work achieves velocities in the 1.5-1.8 m/s and 3.3-5.8 m/s  ranges \cite{margolis2024rapid}, \cite{aractingi2023controlling}, \cite{kumar2021rma}, \cite{hwangbo2019learning}, these results are still below HACL performance. Bandit-based curriculum (UCB, Thompson sampling) also performs worse than HACL in terms of S, velocity, and CoT, with static reward strategies \cite{margolis2024rapid}, \cite{aractingi2023controlling} closely following HACL. Also, we train each robot separately (Mini Cheetah, Unitree Go1 and Go2) on our HACL policy to demonstrate that the policy generalizes regardless of morphological variations and exhibits target curriculum expansion (Fig. 3 and 4).

% In addition to quadrupedal morphological generalization, HACL improves speed, sample efficiency, and stability and exhibits a targeted curriculum expansion (Fig. 3 and 4).

% \subsection{Hardware}

\subsection{Zero-shot Hardware Transfer (Unitree Go1)}
To evaluate sim-to-real robustness, we deploy our trained HACL policy in a zero-shot transfer on a Unitree Go1 EDU robot. The robot has 12 actuated joints, and our policy outputs a joint command of size 12, which the robot executes in the real world. The policy runs on-board compute using a sensor suite of an IMU, force sensors, and encoders.
HACL achieves very rapid and stable locomotion reaching $4.1 \pm 0.2$ m/s, given hardware safety issues and lab limitations, which we have not tested at very high speeds. Previous works such as \cite{margolis2024rapid}, \cite{aractingi2023controlling} report 3.9 m/s on the Mini Cheetah and approximately 1.5 m / s on the Solo12 robot with other RL controllers \cite{hwangbo2019learning}, \cite{kumar2021rma} did not report maximum speed. These findings are consistent with our understanding that a history-aware curriculum significantly improves the speed and stability of locomotion.

\subsection{Terrain Robustness}
To test the robustness of our trained HACL policy, we tested Go1 on rigid, deformable, irregular terrains, and moderate slopes.

\textbf{Rigid-Indoor:} On carpet (reference) policy reaches $4.1 \pm 0.2$ m / s with a success rate 90\%. The tile introduces occasional slips / foot glitches, with a success rate of 100\% and a speed of $3.1 \pm 0.4$ m/s ($\approx 24.0\%$ below the carpet).

\textbf{Rigid-Outdoor:} On cement floor, the performance exceeds the tile: $3.3 \pm 0.3$ m / s with a 80\% success rate (20.0\% below the carpet).

\textbf{Deformable Terrain:} On grass surface, policy performs worse because these surfaces are uneven and deformed, but HACL still achieves a velocity of $1.8 \pm 0.2$ m / s (56.0\% below the carpet).

\textbf{Irregular Terrain:} On pebbles, the task success rate was reduced to 60 %
with $ \approx 2.1 \pm 0.3$ m/s and modest drift $0.3 \pm 0.1$. In broken rocks, the success is 50\% with $ \approx 1.5 \pm 0.4$ m / s and drift $0.3 \pm 0.2$ m (48. 8\% and 63. 4\% below the carpet).

\textbf{Slopes:} Uphill: cement ($15^\circ$) reaches $2.7 \pm 0.3$ (90\%); wood ($20^\circ$) reaches $3.1 \pm 0.4$ m / s (80\%) (see Figures 1 and 6) (34. 1\% and 24. 0\% below the carpet, respectively).

\begin{figure}[h]
    \centering

     % Second row of images
    \begin{minipage}{0.32\linewidth}
        \centering
        \includegraphics[width=\linewidth]{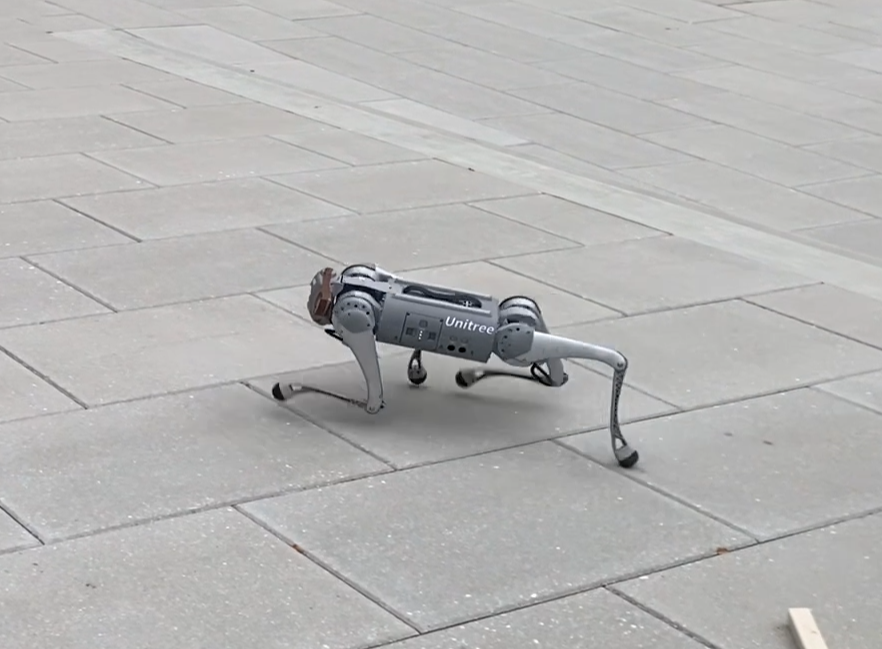}
    \end{minipage}
    \begin{minipage}{0.32\linewidth}
        \centering
        \includegraphics[width=\linewidth]{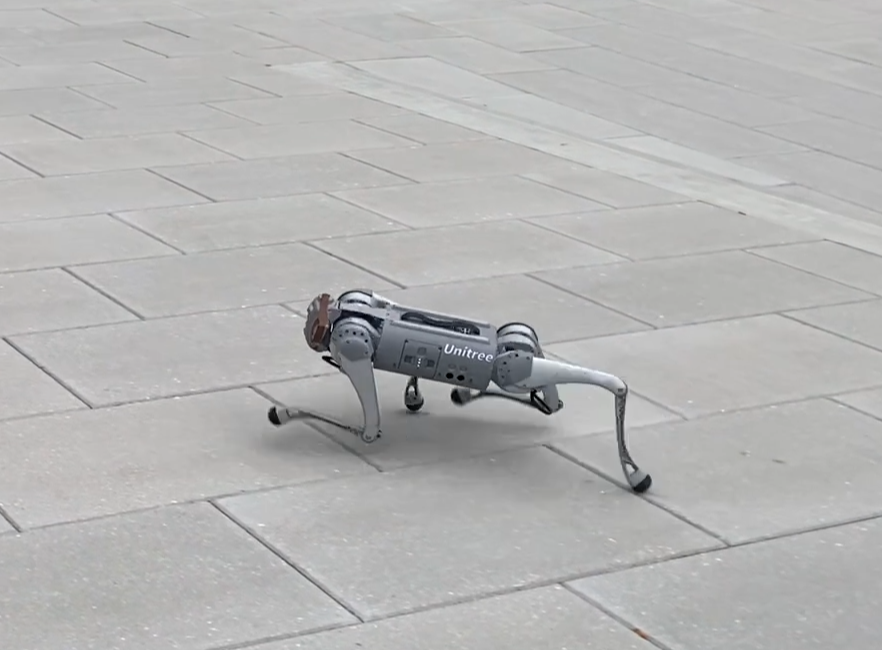}
    \end{minipage}
    \begin{minipage}{0.32\linewidth}
        \centering
        \includegraphics[width=\linewidth]{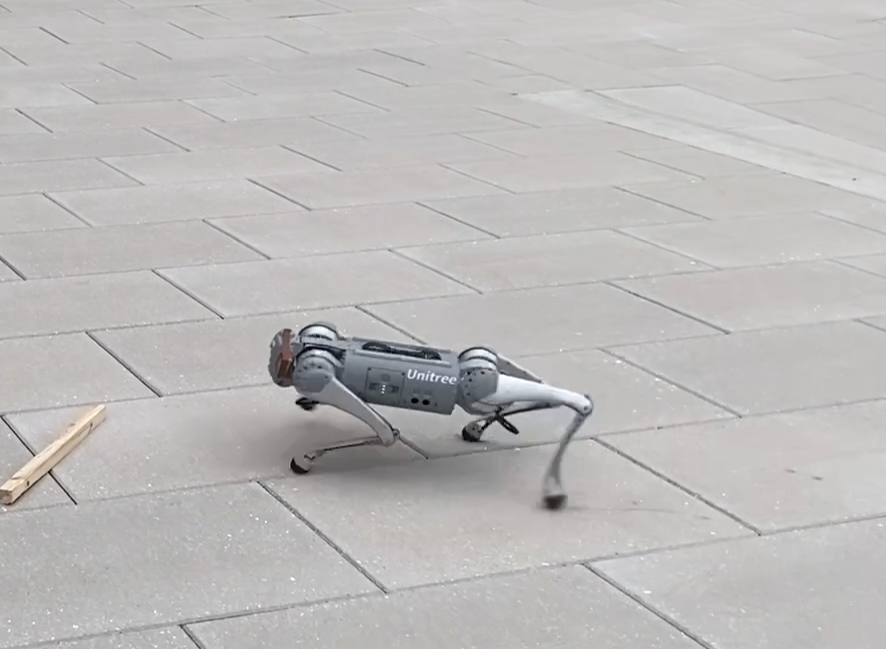}
    \end{minipage}
    
      % Second row of images
    \begin{minipage}{0.32\linewidth}
        \centering
        \includegraphics[width=\linewidth]{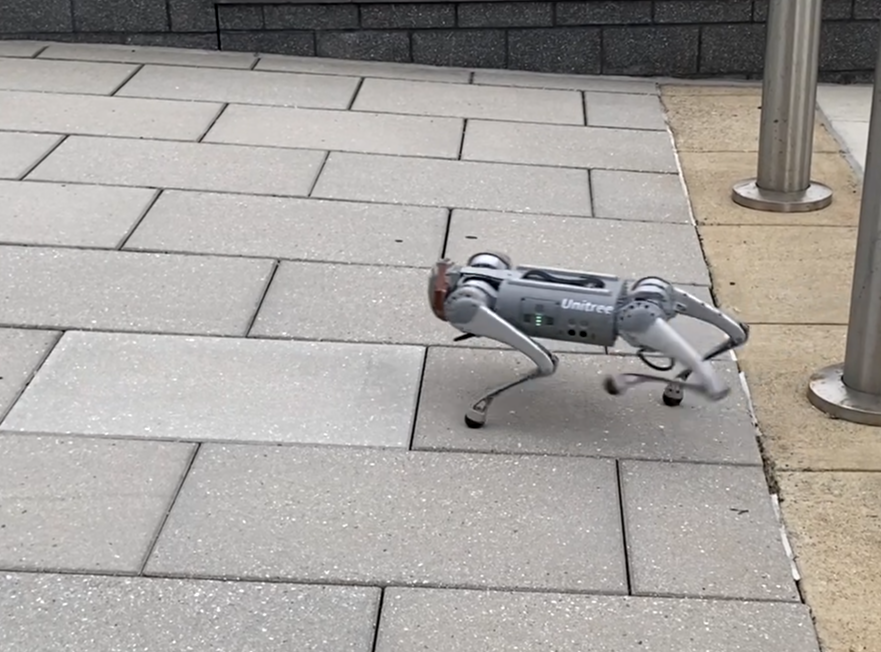}
    \end{minipage}
    \begin{minipage}{0.32\linewidth}
        \centering
        \includegraphics[width=\linewidth]{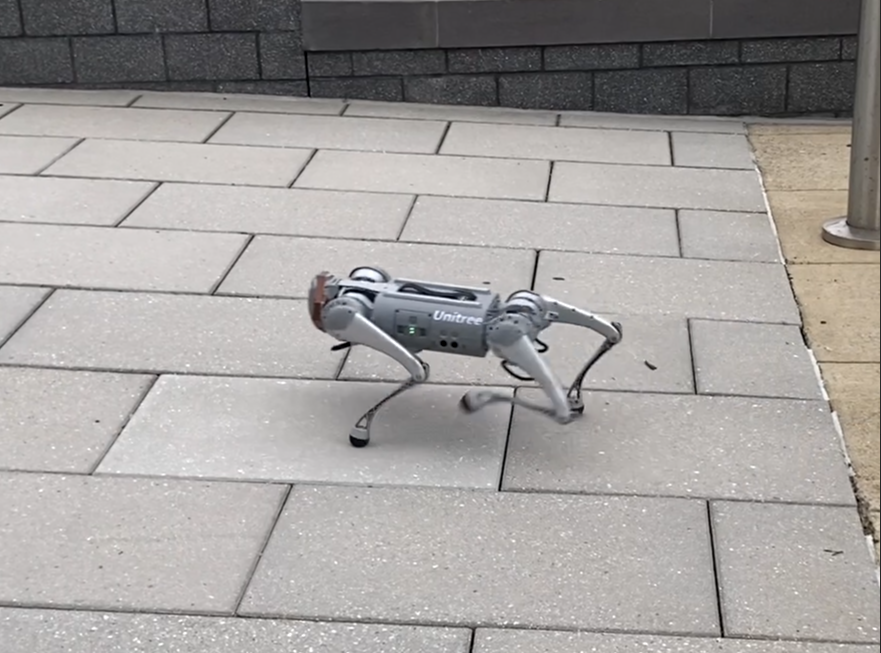}
    \end{minipage}
    \begin{minipage}{0.32\linewidth}
        \centering
        \includegraphics[width=\linewidth]{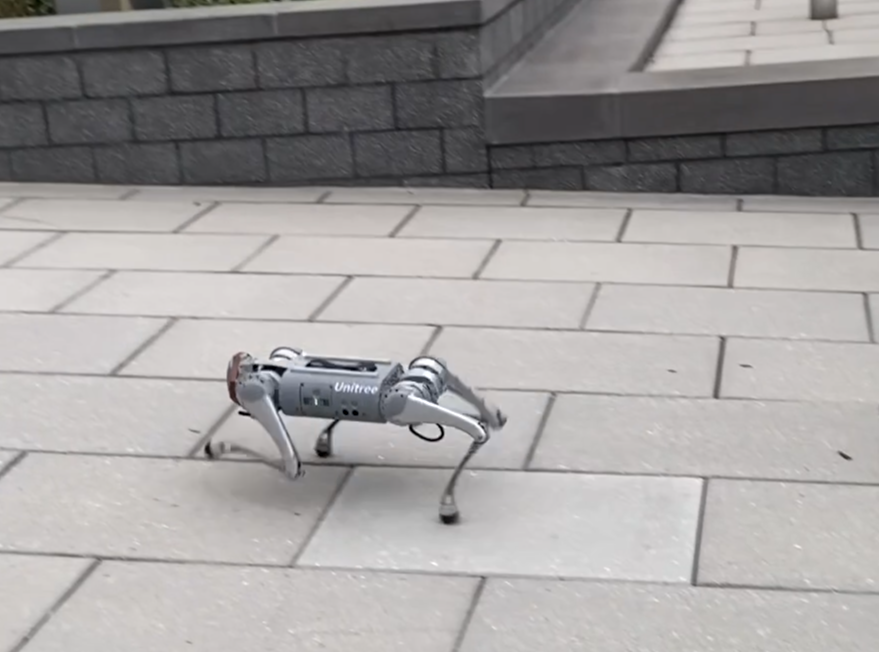}
    \end{minipage}

    %   % Second row of images
    % \begin{minipage}{0.32\linewidth}
    %     \centering
    %     \includegraphics[width=\linewidth]{img/carp1.png}
    % \end{minipage}
    % \begin{minipage}{0.32\linewidth}
    %     \centering
    %     \includegraphics[width=\linewidth]{img/carp2.png}
    % \end{minipage}
    % \begin{minipage}{0.32\linewidth}
    %     \centering
    %     \includegraphics[width=\linewidth]{img/carp3.png}
    % \end{minipage}

      % Second row of images
    \begin{minipage}{0.32\linewidth}
        \centering
        \includegraphics[width=\linewidth]{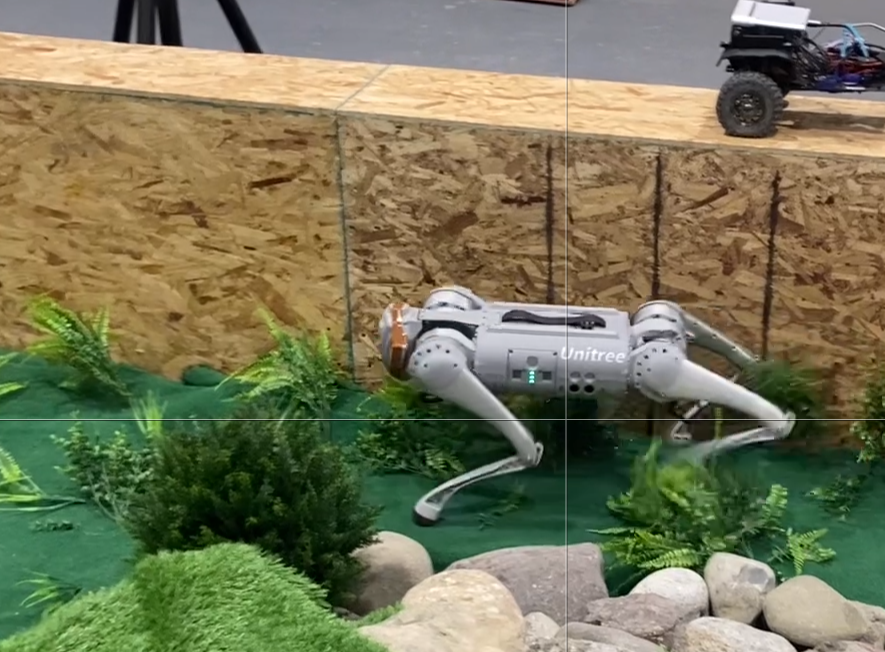}
    \end{minipage}
    \begin{minipage}{0.32\linewidth}
        \centering
        \includegraphics[width=\linewidth]{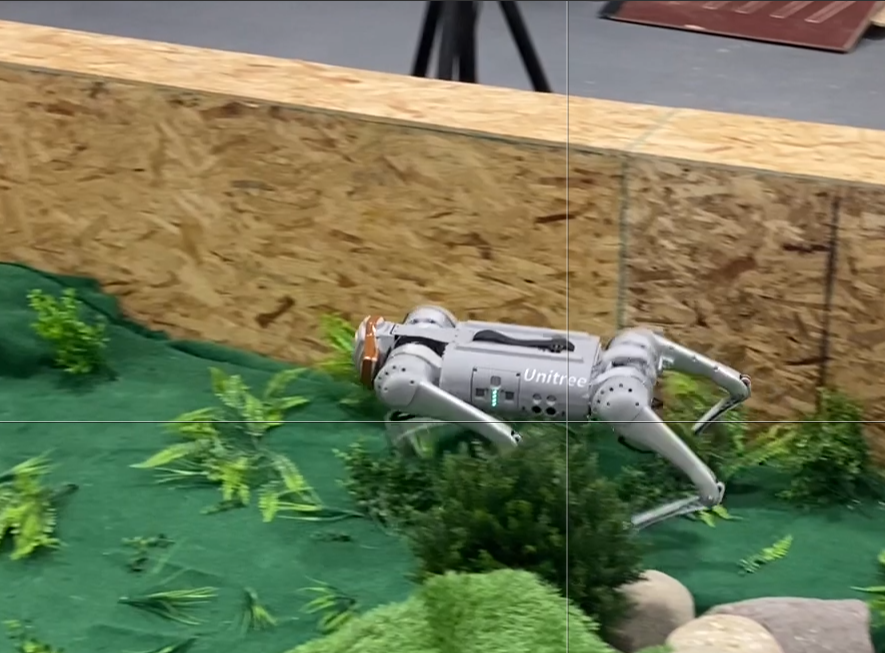}
    \end{minipage}
    \begin{minipage}{0.32\linewidth}
        \centering
        \includegraphics[width=\linewidth]{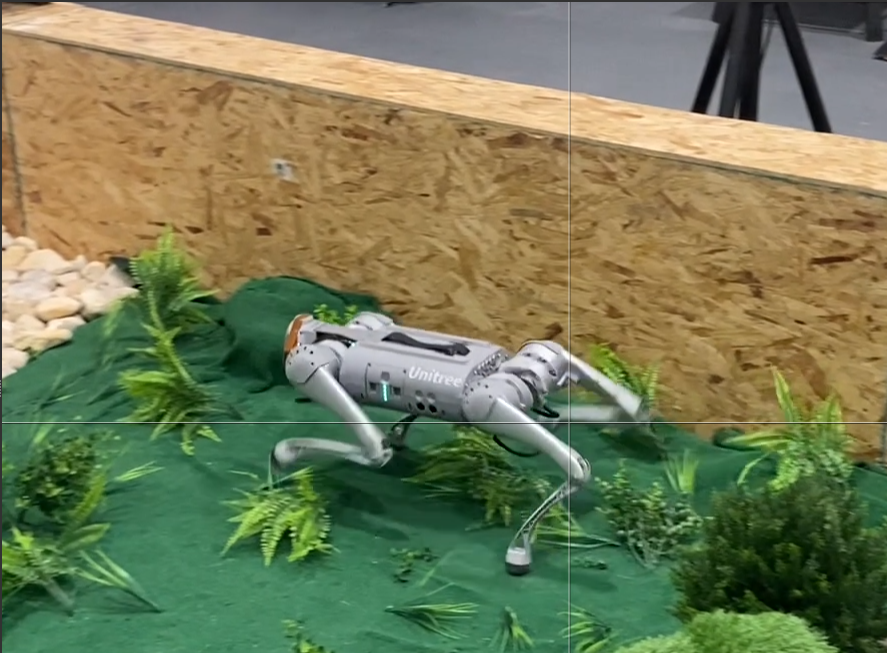}
    \end{minipage}

    %  % Second row of images
    % \begin{minipage}{0.32\linewidth}
    %     \centering
    %     \includegraphics[width=\linewidth]{img/dt1.png}
    % \end{minipage}
    % \begin{minipage}{0.32\linewidth}
    %     \centering
    %     \includegraphics[width=\linewidth]{img/dt2.png}
    % \end{minipage}
    % \begin{minipage}{0.32\linewidth}
    %     \centering
    %     \includegraphics[width=\linewidth]{img/dt3.png}
    % \end{minipage}

     % Second row of images
    \begin{minipage}{0.32\linewidth}
        \centering
        \includegraphics[width=\linewidth]{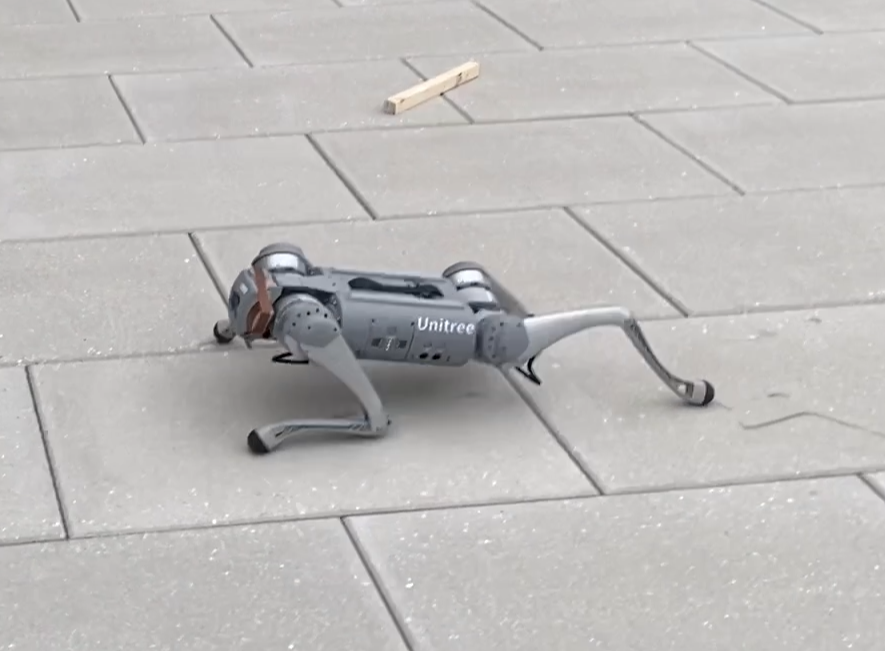}
    \end{minipage}
    \begin{minipage}{0.32\linewidth}
        \centering
        \includegraphics[width=\linewidth]{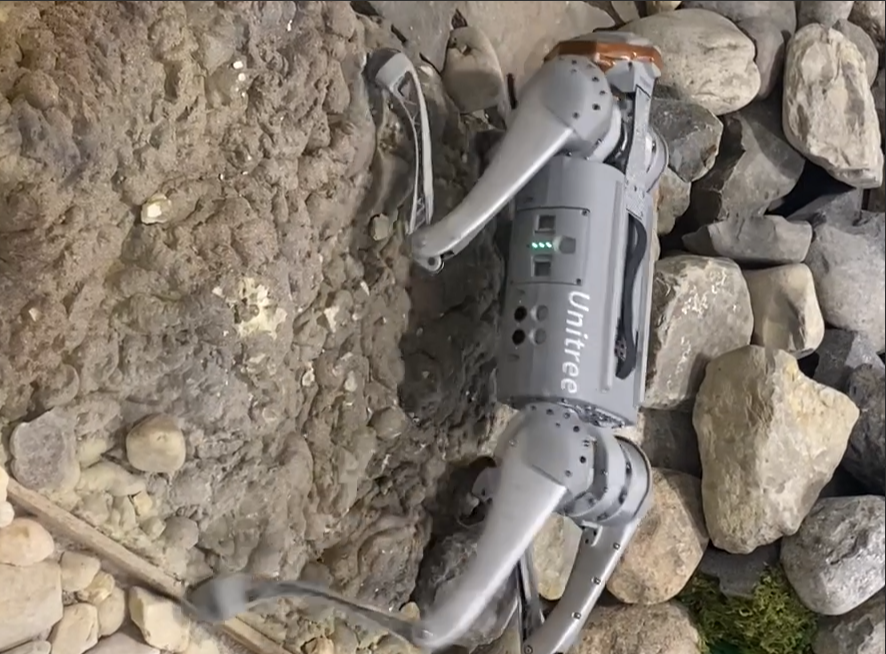}
    \end{minipage}
    \begin{minipage}{0.32\linewidth}
        \centering
        \includegraphics[width=\linewidth]{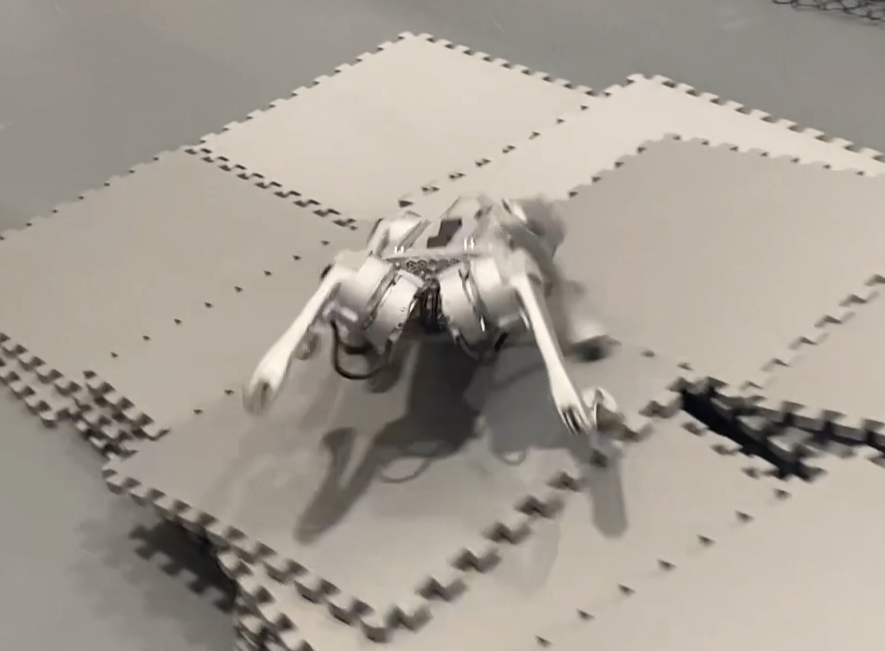}
    \end{minipage}
          
    \caption{\textbf{Agility and robustness on diverse terrain (Unitree Go1):} Each panel is terrain/slope, length of run, achieved velocity, and success rate. 
      \textbf{Rows 1-3 (Terrains):} Cement (8-15\,m, $3.3\pm0.3$\,m/s, $80\%$ success), Slope cement (15 degrees)(3-5\,m, $2.7\pm0.3$\,m/s, $90\%$ success), Grass (2-4\,m, $1.8\pm0.2$\,m/s, $70\%$ success).
    \textbf{Row 4 (Deviation \& Recovery):} Under bumps and trips/falls, our policy re-stabilizes within milliseconds and returns to a stable gait. (Refer Video)}
    \label{fig:hardware_robust}
\end{figure}

The key takeaway from our terrain testing is that, depending on the terrain compliance and irregularity, the performance degrades in the form of energy loss, which shows up as reduced speed and lateral deviations. However, our HACL policy remains functional without any need for fine-tuning, indicating the robustness of the history-aware curriculum.

\subsection{Ablation Studies}
We have compared the history-aware (RNN/LSTM/GRU) curriculum with non-history (MLP/CNN) curriculum under similar training setup including learning rates, bin size, PPO parameters, etc. History matters, as it ($h_{t-1}$) yields efficient and adaptive learning and significantly improves forward velocity, energy efficiency, stability, and task success rate. 

At a command velocity of 7 m/s, history-aware networks (RNN/LSTM/GRU) reach 6.58-6.72 m/s, while non-history networks (CNN/MLP) linger around 0.5-0.6 m/s. Recurrent curricula have a 90\% task success rate and lower CoT compared to a non-history curriculum, which has a 0\% task success rate and higher CoT (50-100 x recurrent ones).

% History really matters as history-aware networks (RNN/LSTM/GRU) yields ~ 10 time higher velocity compared to non-history networks (CNN/MLP) and 3-4 times better energy efficiency with roughly 90\% task success rate, and the desired outcome gets worse at higher command velocity ($v^{cmd}_x$ = 7). While non-history networks (CNN/MLP) achieves only crawling velocity of around ~ 0.6 m/s.

Across RNN/LSTM/GRU, performance differences in terms of final velocity, rewards, and success rate are minor, indicating that the memory itself, instead of recurrent cell architecture, drives improvement with slight variations in CoT and stability. These observations substantiate the importance of history in curriculum-based locomotion learning.

% \begin{table}[t]
% \centering
% \caption{Real-world HACL Policy testing on the Go1 robot(\(n=10\) runs per terrain).}
% \label{tab:surfaces}
% \scriptsize % or \footnotesize
% \setlength{\tabcolsep}{3pt} % default is ~6pt
% \renewcommand{\arraystretch}{0.9} % shrink row height
% \begin{tabular}{lccccc}
% \toprule
% \textbf{Terrain} &
% \textbf{Distance(m)}  &
% \textbf{Success \%age}  &
% \textbf{$V_{max}$}  &
% \textbf{Lateral Shift} \\
% \midrule
% Cement & $8 - 15 $ & 80  & $3.3 \pm 0.3$ & $0.9 \pm 0.3$\\
% Woolen carpet     & $3 - 8 $ & 90  &  $4.1 \pm 0.2$ & $1.1 \pm 0.4$\\
% Tile        & $3 - 10 $ & 100  & $3.1 \pm 0.4$ & $1.2 \pm 0.2$\\
% Slope cement ($15^\circ$)       & $3 - 5 $ & 90  &  $2.7 \pm 0.3$ & $0.6 \pm 0.2$\\
% Slope wooden ($20^\circ$)      & $3 - 5 $ & 80  &  $3.1 \pm 0.4$ & $0.5 \pm 0.3$\\
% Grass         & $2 - 4 $ & 70  &  $1.8 \pm 0.2$ & $0.4 \pm 0.2$\\
% Pebbles         & $2 - 3 $ & 60  &  $2.1 \pm 0.3$ & $0.3 \pm 0.1$\\
% Broken Rocks         & $2 - 3 $ & 50  &  $1.5 \pm 0.4$ & $0.3 \pm 0.2$\\
% \bottomrule
% \end{tabular}
% % \vspace{-3mm}
% \vspace{2pt}
% % \footnotesize All the success \%age is reported with (95\% CI).
% \end{table}

\subsection{Failure Modes and discussion}

We observe three recurrent modes of failure on the Unitree Go1 robot (i) lateral deviation while running in a straight path of 2-8 m, (ii) slips/micro-bumps on low friction and irregular terrains, (iii) and transient gait irregularities where the robot goes from trot to crawl (0.5-1.0 s) due to misaligned contact.

\begin{table}[h]
 \caption{{History vs. non-history methods (\(n=5\) runs per method)}}
\centering

\resizebox{\columnwidth}{!}{%
\begin{tabular}{lccccc}
\hline
\textbf{Metric}                  & \textbf{RNN} & \textbf{LSTM} & \textbf{GRU} & \textbf{CNN} & \textbf{MLP} \\ \hline

$v_x$ m/s ($v^{cmd}_x = 6$)      & 6.0          & 6.0            & 6.0           & 0.6          & 0.56          \\

$v_x$ m/s ($v^{cmd}_x = 7$)      & 6.62          & 6.58           & 6.72          & 0.62          & 0.5         \\

$\omega_z$ rad/s   & 1.25          & 1.2           & 1.28          & 0          & 0         \\
$r_{lin}$   & 0.0823          & 0.0873            & 0.085          & 0.08          & 0.07          \\
History-Aware          & Yes          & Yes           & Yes          & No           & No           \\
CoT           & 2.60           & 4.00            & 5.20          & 60 x RNN           & 100 x RNN           \\

Stability (S)                       & 2000         & 2100        & 1900         & 2000       & 1100       \\ 
Task Success Rate                & 90\%         & 90\%          & 90\%           & 0\%       & 0\%       \\  \hline
\end{tabular}%
}
   
\label{tab:history_vs_non}
\end{table}

Failure modes drive a terrain-dependent gap in success and speed for e.g., tile, woolen carpet, and cement slopes have high task success rate ranging from 90- 100\%, while it drops by (20-30\%) on grass, pebbles, and broken stone terrain. The maximum forward velocity in tile, wool carpet, and cement ranges from 3.0-4.1 m /s and drops significantly to 1.8-2.1 m / s for grassy terrain, pebbles, and broken stones.

Likely causes include (i) reward shaping focused on speed and stability rather than running in straight paths, yielding lateral deviation and some friction/irregularity of terrain, (ii) substrate mismatch e.g., deformable/loose surfaces (like pebbles, tile, broken rocks) that were not modeled in the simulator, reducing traction and causing slips, and (iii) contact inference failure causing short gait dropout. However, the policy still performed well on such unseen terrains and adapted its gait quickly. 

The sim-to-real gap is consistent; the policy reaches 6.7 m/s in simulation at the commanded velocity but degrades to $4.1 \pm 0.2$ m/s in the real world. We attribute this gap to mainly (i) end-to-end actuator/sensor latency, (ii) unmodeled terrain parameters beyond our DR limits, (iii) state-estimation delay causing conservative policy output, and (iv) hardware specific drop due to joints heat-up. In general, HACL performs exceptionally well on rigid surfaces such as tiles, wool carpet, and cement and recovers quickly from bumps and slips, remaining functional on irregular terrains and bumpy slopes without any fine-tuning (Figure 6).

\section{Conclusion, Limitations, and Future Work}

HACL enables fast and stable quadrupedal locomotion by leveraging temporal dependencies using a history-based curriculum. The benefits of our approach include: $(i)$ higher linear velocity; $(ii)$ better stability than contemporary methods~\cite{margolis2024rapid}, \cite{kumar2021rma}, \cite{aractingi2023controlling}; $(iii)$ higher task success rate; $(iv)$ higher learning signal ($\uparrow r_{\text{lin}}$); and $(v)$  lower energy consumption \((\downarrow CoT)\), even at higher command velocity. 
Ablations show that ignoring history at the curriculum level reduces speed and stability and prevents a robot from achieving its full locomotion potential. HACL generalizes across different robot morphologies (Mini Cheetah, Unitree Go1, Go2), one HACL policy per robot without any morphological fine-tuning. Some of the key limitations of our policy are that we have not explicitly conditioned on robot morphologies \cite{huang2020one} and also not tested it for bipedal/humanoid robots, which we leave it for future work. Evaluations of Go1 on various terrains and slopes demonstrate the robustness and utility of HACL's history-aware curricula in the context of locomotion.

% \cite{pathak2019learning}, morphology 

%                                   % the textheight of the last page by a suitable amount.
%                                   % This command does not take effect until the next page
%                                   % so it should come on the page before the last. Make
%                                   % sure that you do not shorten the textheight too much.

% %%%%%%%%%%%%%%%%%%%%%%%%%%%%%%%%%%%%%%%%%%%%%%%%%%%%%%%%%%%%%%%%%%%%%%%%%%%%%%%%

% %%%%%%%%%%%%%%%%%%%%%%%%%%%%%%%%%%%%%%%%%%%%%%%%%%%%%%%%%%%%%%%%%%%%%%%%%%%%%%%%

% %%%%%%%%%%%%%%%%%%%%%%%%%%%%%%%%%%%%%%%%%%%%%%%%%%%%%%%%%%%%%%%%%%%%%%%%%%%%%%%%
% % \section*{APPENDIX}

% % Appendixes should appear before the acknowledgment.

% % \section*{ACKNOWLEDGMENT}

% % The preferred spelling of the word ÒacknowledgmentÓ in America is without an ÒeÓ after the ÒgÓ. 

% % \begin{thebibliography}{99}

% % \bibitem{c1}
% % G. Margolis, G. Yang, K. Paigwar, T. Chen, and P. Agrawal,
% % ``Rapid locomotion via reinforcement learning,''
% % \textit{The International Journal of Robotics Research}, vol. 43, pp. 572--587, 2024.

% \bibliographystyle{IEEEtran}
% \bibliography{refs}  
% % \end{thebibliography}

% \end{document}

\balance
% (Optional, fine-grained control instead of \balance:)
% \IEEEtriggeratref{22}
% \IEEEtriggercmd{\enlargethispage{-2.0in}}

% \section*{References}
\bibliographystyle{IEEEtran}
\bibliography{refs}

\end{document}